\documentclass[10pt]{article} 
\usepackage[preprint]{tmlr}

\usepackage{amsmath,amsfonts,bm}

\def\eqref#1{equation~\ref{#1}}

\def\1{\bm{1}}

\DeclareMathAlphabet{\mathsfit}{\encodingdefault}{\sfdefault}{m}{sl}
\SetMathAlphabet{\mathsfit}{bold}{\encodingdefault}{\sfdefault}{bx}{n}

\usepackage{hyperref}
\usepackage{url}
\usepackage{booktabs}       
\usepackage{amsfonts}       
\usepackage{nicefrac}       
\usepackage{microtype}      
\usepackage[dvipsnames]{xcolor} 
\usepackage{lipsum} 
\usepackage{bbm}
\usepackage{multirow}
\usepackage{relsize}
\usepackage{array}
\usepackage{colortbl}
\usepackage{pifont}
\usepackage{hhline}
\usepackage{boldline}
\usepackage{float}
\usepackage{pifont}
\usepackage{subcaption}
\usepackage[figuresright]{rotating}
\usepackage{balance}
\usepackage{blindtext}
\usepackage{paralist}
\usepackage{arydshln} 
\usepackage{wrapfig}
\usepackage{colortbl}
\usepackage[most]{tcolorbox}
\tcbuselibrary{listings,breakable}
\usepackage{fontawesome5}
\usepackage[inline]{enumitem}
\usepackage{soul}
\sethlcolor{orange!10}

\usepackage{xspace}
\usepackage{setspace}
\newcommand{\ourmethod}{{\fontfamily{lmtt}\selectfont \textbf{RoboEvolve}}\xspace}

\newcommand{\obsbox}[1]{%
    \begin{tcolorbox}[colframe=black!70, colback=cyan!2, boxrule=1pt, arc=2mm,   top=2pt, bottom=3pt, left=3pt, right=3pt,
  boxsep=1pt]
        #1
    \end{tcolorbox}
}

\newcommand{\emailicon}{{\fontsize{8}{10}\selectfont\faEnvelope}}

\hypersetup{
    colorlinks=true,
    linkcolor=RoyalBlue,
    citecolor=RoyalBlue,
    filecolor=RoyalBlue,      
    urlcolor=RoyalBlue,
    }

\title{RoboEvolve: Co-Evolving Planner-Simulator for Robotic\\Manipulation with Limited Data}

\author{Harold Haodong Chen$^{1\dagger}$,
   Sirui Chen$^{2\dagger}$,
   Yingjie Xu$^{2}$,
   Wenhang Ge$^{2}$,
   Ying-Cong Chen$^{2,1*}$ \\
   \newline
   $^{1}$The Hong Kong University of Science and Technology\\
   $^{2}$The Hong Kong University of Science and Technology (Guangzhou)\\
   \newline
   \emailicon\ haroldchen328@gmail.com, \quad
   $^{\dagger}$\textit{Equal Contribution}, \quad
   $^{*}$\textit{Corresponding Author}
}

\def\month{MM}  
\def\year{YYYY} 
\def\openreview{\url{https://openreview.net/forum?id=XXXX}} 

\begin{document}

\maketitle

\begin{abstract}
The scalability of robotic manipulation is fundamentally bottlenecked by the scarcity of task-aligned physical interaction data. While vision-language models (VLMs) and video generation models (VGMs) hold promise for autonomous data synthesis, they suffer from semantic-spatial misalignment and physical hallucinations, respectively. To bridge this gap, we introduce \ourmethod, a novel framework that couples a VLM planner and a VGM simulator into a mutually reinforcing co-evolutionary loop. Bootstrapped for self-evolution with \textit{unlabeled seed images}, \ourmethod leverages a cognitive-inspired dual-phase mechanism: (\textbf{\textit{i}}) \textbf{daytime exploration} fosters physically grounded behavioral discovery through a semantic-controlled multi-granular reward, and (\textbf{\textit{ii}}) \textbf{nighttime consolidation} mines "near-miss" failures to stabilize policy optimization. Guided by an autonomous progressive curriculum, the system naturally scales from simple atomic actions to complex tasks. Extensive experiments demonstrate that \ourmethod \textbf{(I)} achieves superior effectiveness, elevating base planners by $30$ absolute points and amplifying simulator success by $48\%$ on average; \textbf{(II)} exhibits extreme data efficiency, surpassing fully supervised baselines with merely $500$ unlabeled seeds--a $50\times$ reduction in initial RL data; \textbf{(III)} demonstrates robust continual learning without catastrophic forgetting; and \textbf{(IV)} extends to low-level control, improving cold-started LingBot-VA from $78.7\%$ to $98.7\%$ average task success on LIBERO using environment feedback. Together, these results demonstrate the applicability of our co-evolutionary framework to both internally supervised generative learning and environment-grounded action learning.
\end{abstract}

\section{Introduction}
\label{sec:intro}

The transition from digital intelligence to physical intelligence represents one of the most profound challenges of today. Although foundation models \citep{achiam2023gpt, team2023gemini, wan2025wan, bai2023qwen, sora2_openai_2025} have significantly advanced semantic understanding across vision and language domains, transferring these capabilities to embodied robotic manipulation remains constrained by a fundamental bottleneck: \textit{the lack of scalable, task-aligned interactive data and supervision}.

\begin{figure*}[!t]
\centering
\includegraphics[width=\linewidth]{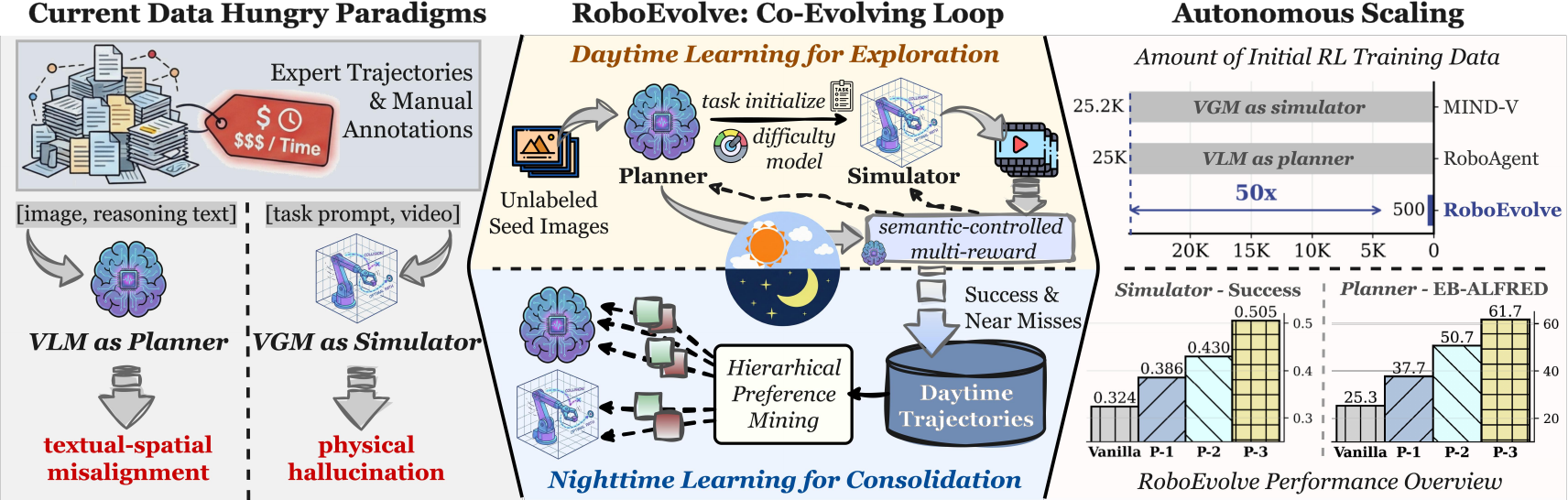}
\vspace{-1.6em}
\caption{(\textbf{\textit{Left}}) Current data-hungry paradigms for robotic manipulation. (\textbf{\textit{Middle}}) \ourmethod's co-evolving loop overview. (\textbf{\textit{Right}}) \ourmethod achieves a $50\times$ reduction in initial RL training data while maintaining monotonic performance gains across iterative day-night phases (P-1 to P-3).}
\label{fig:figure1}
\vspace{-0.6em}
\end{figure*}

High-quality robot trajectories are notoriously expensive and time-consuming to collect, especially when they require precise annotations or human demonstrations \citep{bai2025towards, shao2025large}. This scarcity of data creates a critical barrier to progress in robotic manipulation. To address this, researchers have turned to two emerging paradigms (see Figure~\ref{fig:figure1} (\textit{Left})): \textbf{(I) vision-language models (VLMs)} \citep{team2023gemini, achiam2023gpt, bai2023qwen, guo2025seed1, dong2024internlm} excel at semantic scene understanding and can generate high-level plans, making them attractive candidates as the "brain" of embodied agents \citep{fang2025robix, team2025robobrain, ji2025robobrain, agarwal2025cosmos}. However, their plans often inevitably lack grounding in physical realities, as their internalization of spatial-physical reasoning within a textual space \citep{he2025vision, park2025making}, making scaling VLMs for manipulation requires robust verification to ensure plan feasibility, which remains impractical without extensive manual supervision.
In parallel, \textbf{(II) video generation models (VGMs)} \citep{wan2025wan, kong2024hunyuanvideo, yang2024cogvideox, sora2_openai_2025} offer the potential to synthesize large-scale interaction data, providing a scalable alternative to labor-intensive robot trajectory collection \citep{zhang2025mind, fu2025learning, chi2025wow, zhou2024robodreamer}. However, owing to the scarcity of task-aligned interaction data for training, VGMs also often suffer from physical hallucination, producing visually plausible but physically infeasible trajectories that fail to achieve the intended task goals, limiting their utility for embodied learning \citep{mei2026video, ding2025understanding}.

Given these limitations, we advocate a hypothesis that \textit{VLMs and VGMs can mutually assist each other in robotic manipulation tasks}. Specifically, VLMs can provide diverse task prompts and judgments that guide VGMs toward more meaningful and semantically grounded trajectory generation, while VGMs simulate the physical feasibility of tasks and provide critical feedback to refine VLM planning. However, to the best of our knowledge, \textbf{no} prior work has explored this problem directly. Related efforts have largely focused on either VLM/LLM self-play evolution \citep{huang2025r, zhao2025absolute, he2025visplay} or VGM-based reinforcement learning (RL) with VLM-based rewards \citep{zhang2025mind}. Yet, we also observe a critical gap that they overwhelmingly focus on successful trajectories during online RL while neglecting the valuable insights that can be extracted from failure cases, making direct transfer of such ideas still inefficient. 
These aforementioned observations bring us to our pivotal research question:

\vspace{-0.4em}
\obsbox{
\includegraphics[height=0.9em]{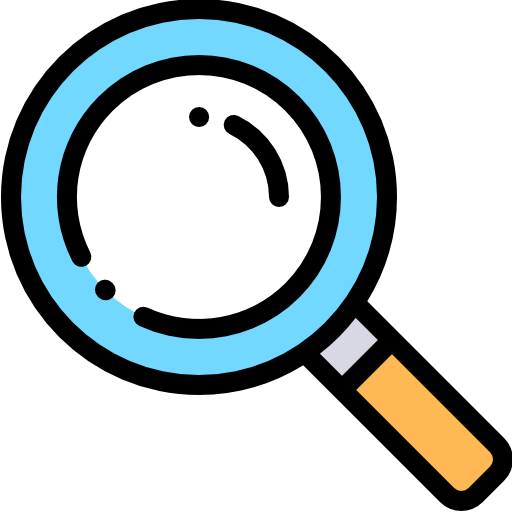}~
\textit{How can we design a collaboration system that couples a VLM planner and a VGM simulator to supervise and evolve each other, effectively leveraging both successes and failures, while scaling from limited, unlabeled data?}}
\vspace{-0.1em}

To bridge this gap, we propose \ourmethod, a novel self-evolving framework for robotic manipulation that integrates a \ding{168} \textbf{planner} (VLM) and a \ding{171} \textbf{simulator} (VGM) into a co-evolving system. Inspired by the Complementary Learning Systems (CLS) theory \citep{mcclelland1995there, kumaran2016learning} in cognitive science, which posits that effective learning emerges from the interplay between exploratory and consolidative processes, \ourmethod operates through a dual-phase evolution loop, as shown in Figure~\ref{fig:figure1} (\textit{Middle}):

\vspace{-0.4em}
\begin{itemize}[leftmargin=1.6em]
    \item[\faSun] \textbf{Daytime Learning} for online exploration: The planner generates executable tasks based on \textit{scene-grounded initialization}, while the simulator generates and simulates trajectories, with a \textit{semantic-controlled multi-granular reward} mechanism that ensures physical realism and semantic consistency to guide the online RL process.
    \item[\faMoon] \textbf{Nighttime Learning} for offline consolidation: Just as humans consolidate experiences during sleep, \ourmethod systematically mines failure cases from daytime and applies a \textit{hierarchical preference optimization} strategy to refine both the planner and simulator under offline policy, ensuring even unsuccessful attempts contribute to learning.
\end{itemize}
\vspace{-0.4em}

These two phases are interleaved in a continual loop, guided by an \textit{atomic-action difficulty function} that progressively evolves task complexity while preserving executability. Daytime learning provides breadth by generating diverse hypotheses and ensuring extensive behavioral coverage, while nighttime learning offers depth through systematic correction and stabilization via failure analysis.
Together, \ourmethod achieves high data efficiency in its VLM--VGM self-evolution stage, requiring only a small set of unlabeled images without additional human annotations or external reward signals, as shown in Figure~\ref{fig:figure1} (\textit{Right}). The same co-evolutionary design also supports low-level action learning with feedback from a physics-based environment.
To summarize, our contributions are threefold:
\vspace{-0.4em}
\begin{itemize}[leftmargin=1.6em]
    \item[\ding{182}] \textbf{\ourmethod Framework.} We introduce \ourmethod, a novel self-evolving framework that couples a vision-language planner and a video generation simulator. By integrating scene-grounded atomic-action difficulty modeling, \ourmethod enables continual learning from simple to complex manipulation, with its self-evolution stage using unlabeled images without additional external annotations or rewards.
    \item[\ding{183}] \textbf{Dual-Phase Evolution Loop.} We propose a cognitive science-inspired daytime-nighttime evolution loop, where daytime encourages diverse and physically grounded exploration through a semantic-controlled multi-granular reward mechanism, and nighttime consolidates experience by leveraging both successes and failures via hierarchical preference optimization.
    \item[\ding{184}] \textbf{Empirical Evaluation.} Extensive experiments demonstrate that \ourmethod achieves: (\textit{\textbf{i}}) \textbf{superior effectiveness}, amplifying simulator relative success gains by $48\%$ on BridgeData V2 and elevating base planners by $30$ absolute points on EB-ALFRED and EB-Habitat; (\textit{\textbf{ii}}) \textbf{extreme data efficiency}, surpassing fully-supervised baselines using merely $500$ unlabeled seeds (a $50$$\times$ reduction in annotations); and (\textit{\textbf{iii}}) \textbf{robust continual learning}, maintaining monotonic capability improvements across increasingly complex tasks without catastrophic forgetting. We further demonstrate its (\textbf{\textit{iv}}) \textbf{applicability to low-level control}, improving average LIBERO success by $20.0$ absolute points over the cold-started model.
\end{itemize}
\vspace{-0.4em}
\section{Related Work}

\vspace{-0.4em}
\paragraph{Vision-Language Models as Planners.}
The emergent reasoning of VLMs has established them as the "brain" for embodied agents \citep{brown2020language, team2023gemini, achiam2023gpt, bai2023qwen}. Conventional paradigms fine-tune VLMs to map observations into textual instructions \citep{fang2025robix, ji2025robobrain, team2025robobrain, tan2026robobrain, hao2025mimo};  however, relying solely on internalizing complex spatial/physical reasoning within its textual latent space often leads to a semantic-physical misalignment \citep{he2025vision, park2025making, huang2023voxposer}. Consequently, planners may produce logically coherent but physically infeasible trajectories.
Recent vision-language-action (VLA) models \citep{zitkovich2023rt, wen2025dexvla, wen2024diffusion, huang2025graphcot} attempt to bridge this gap by integrating low-level action heads, yet they remain constrained by the scarcity of high-fidelity, visually diverse data and the prohibitive cost of real-world collection \citep{din2025vision, bai2025towards, bai2025embodied, o2024open}. Moreover, the dependence on rigid reward functions often limits their ability to learn from failure.
In contrast, \ourmethod bypasses these constraints by employing a VGM as a dynamic, learnable world simulator. This also allows the planner to proactively visualize and rectify physical misconceptions through synthesized, multi-granular feedback, transforming failure cases into valuable supervisory signals for self-evolution.

\vspace{-0.4em}
\paragraph{Video Generation Models as Simulators.}
VGMs \citep{he2022latent, wan2025wan, sora2_openai_2025, yang2024cogvideox, kong2024hunyuanvideo} have transitioned from visual synthesis toward capturing physical plausibility, positioning them as neural world models. Within embodied AI, VGMs are increasingly utilized as scalable simulators to bypass the high cost of manual data collection \citep{mei2026video, ding2025understanding}.
Current methodologies primarily fall into two paradigms: (\textbf{\textit{i}})~\textbf{trajectory fitting via SFT} \citep{fu2025learning, agarwal2025cosmos, du2023learning, zhu2024irasim, zhou2024robodreamer}, where VGMs are trained on expert demonstrations but remain bottlenecked by the scarcity of high-quality labels; and (\textbf{\textit{ii}})~\textbf{exploration via RL} \citep{zhang2025mind, guo2025deepseek}, where VGMs serve as interactive environments for policy training. While RL-based methods can theoretically uncover deeper physical insights, they still also depend heavily on pre-annotated, task-specific datasets \citep{ebert2021bridge}, limiting scalability in scenarios with sparse or unlabeled data.
Distinct from these static or data-hungry paradigms, \ourmethod introduces a co-evolving loop. Instead of treating the VGM as a fixed oracle, we leverage a VLM planner to provide semantic anchoring, enabling the VGM to evolve into a task-aligned simulator even from sparse seed images.

\vspace{-0.4em}
\paragraph{Self-Evolving System.}
The concept of self-evolution has recently emerged as a pivotal mechanism to endow models with lifelong learning capabilities \citep{gao2025survey, fang2025comprehensive}. Existing works primarily root in language models, generally follow two paradigms: (\textbf{\textit{i}})~\textbf{experience accumulation} \citep{zhao2024expel, song2024agentbank, zheng2025skillweaver, suzgun2025dynamic, zhang2025darwin}, where models aggregate reasoning trajectories/chains to contextually enhance their future problem-solving skills; and (\textbf{\textit{ii}})~\textbf{self-play \& discovery} \citep{zhao2025absolute, he2025visplay, huang2025r, yue2026dr}, characterized by models autonomously generating challenges and refining their internal policies through active exploration. While our \ourmethod aligns with the self-play paradigm, existing frameworks are almost solely focused on language domains. 
Furthermore, a prevalent limitation in existing systems is their heavy bias toward successful outcomes, often discarding failure cases as non-informative noise. 
Inspired by CLS theory \citep{mcclelland1995there, kumaran2016learning}, \ourmethod extends self-evolution to the embodied domain. Unlike prior success-oriented approaches, we systematically mine failures during a "nighttime learning" phase to refine the system, which ensures that even unsuccessful attempts contribute to the system's consolidation.
\section{Preliminary}
\vspace{-0.6em}

\paragraph{Problem Formulation.}
Our goal is to empower a robotic agent to learn complex manipulation skills from a limited set of unlabeled seed images, denoted as $\mathcal{D} = \{I_1, I_2, \dots, I_N\}$. Each manipulation task is defined as a state transition from an initial state $I$ to a goal state $G$, achieved via a trajectory $\tau$. In our \ourmethod, $\tau$ is represented as a video sequence $V = \{f_1, f_2, \dots, f_T\}$, where each frame $f_t$ corresponds to an intermediate state. This trajectory is synthesized by a video generation model (VGM), acting as a \textbf{simulator} $\mathcal{S}$, conditioned on a plan $\pi$ generated by a vision-language model (VLM) \textbf{planner} $\mathcal{P}$.

Unlike traditional paradigms \citep{zhou2024robodreamer, zhang2025mind, fu2025learning} that rely on predefined simulators or extensive manual annotations, \ourmethod operates in a \textbf{self-evolving environment}. The core objective is to co-evolve the planner $\mathcal{P}$ and the simulator $\mathcal{S}$ in a closed-loop system, such that $\mathcal{P}$ generates physically feasible plans and $\mathcal{S}$ produces high-fidelity, physically consistent simulations, even in the absence of expert demonstrations or ground-truth reward functions. 

\vspace{-0.4em}
\paragraph{Atomic Action and Difficulty Space.}
To bridge the semantic gap between high-level reasoning and low-level execution, we first define an \textbf{atomic action space} $\mathcal{A}$. A plan $\pi$ is decomposed into a sequence of atomic actions $\pi = \langle a_1, a_2, \dots, a_n \rangle$, where each $a_i \in \mathcal{A}$ (\textit{e.g.},{\small\texttt{"pick(X)"}}, {\small\texttt{"place(X, target)"}}) corresponds to a visually identifiable motion segment in the generated video $V$. These atomic actions serve as the fundamental building blocks for constructing complex manipulation tasks, enabling precise alignment between the planner $\mathcal{P}$'s outputs and the simulator $\mathcal{S}$'s execution.

To quantify task complexity, we further introduce a \textbf{difficulty function} $D(\tau|I)$, which evaluates the execution cost of a task $\tau$ given the initial scene $I$:
\setlength\abovedisplayskip{3pt}
\setlength\belowdisplayskip{3pt}
\begin{equation}
    D(\tau|I) = \sum_{a_i \in \pi} c(a_i),
    \label{eq:difficulty}
\end{equation}
where $c(a_i)$ represents the unit cost associated with each atomic action $a_i$. Unlike prior works that rely on static, fixed datasets, this difficulty metric serves as the state variable for \ourmethod's curriculum evolution, guiding the system from simple single-stage manipulations to complex, multi-stage tasks.

\vspace{-0.4em}
\paragraph{Complementary Learning System.}
The evolution mechanism of \ourmethod draws inspiration from the CLS theory \citep{mcclelland1995there, kumaran2016learning}, which interleaves two phases to decouple exploration and consolidation:

\vspace{-0.6em}
\begin{itemize}[leftmargin=1.6em]
    \item[\faSun] \textit{Daytime Exploration}: Analogous to the hippocampal mechanism, the agent performs active exploration. We formulate this as a Group Relative Policy Optimization (GRPO) \citep{guo2025deepseek} process, where groups of plans $\{\pi_1, \dots, \pi_K\}$ or trajectories $\{\tau_1, \dots, \tau_K\}$ are sampled and evaluated to identify relative advantages, fostering discovery and breadth.
    \item[\faMoon] \textit{Nighttime Consolidation}: Inspired by the neocortical process, the agent reviews experiences. We model this as a Direct Preference Optimization (DPO) \citep{rafailov2023direct} process, where preference pairs $(\pi_\text{win}, \pi_\text{lose})$ or $(\tau_\text{win}, \tau_\text{lose})$ are constructed from the successes and failures of the daytime phase, mitigating physical hallucinations in $\mathcal{S}$ and logical fallacies in $\mathcal{P}$.
\end{itemize}
\vspace{-0.6em}

\begin{figure*}[!t]
\centering
\includegraphics[width=\linewidth]{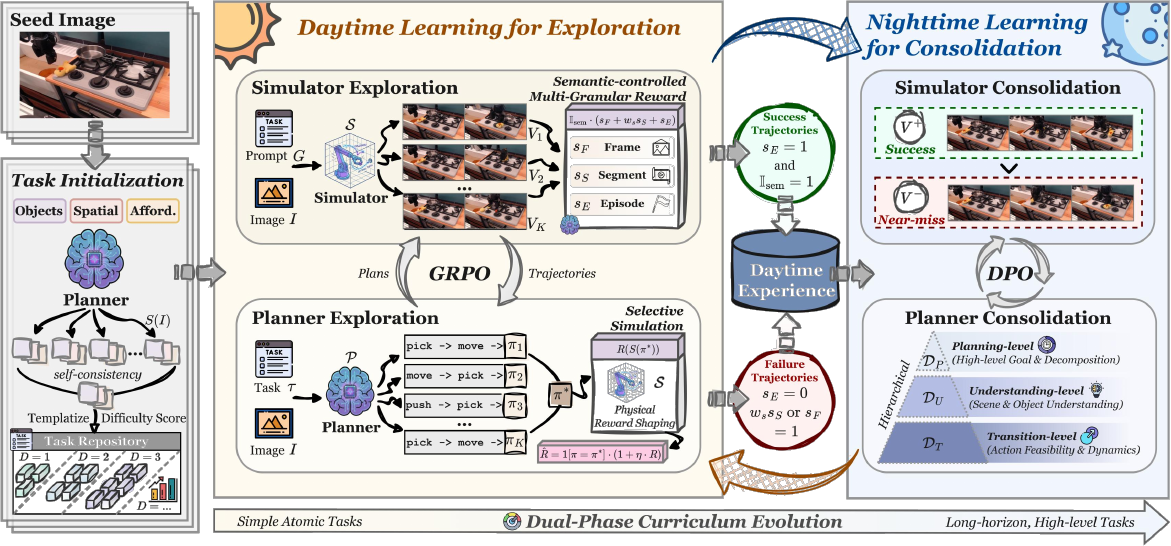}
\vspace{-1.8em}
\caption{Overview of our proposed \ourmethod.}
\label{fig:figure2}
\vspace{-1.3em}
\end{figure*}

\section{Methodology}
\label{sec:method}

\ourmethod establishes a self-evolving loop that interleaves autonomous discovery with knowledge consolidation to bridge the gap between semantic planning and physical execution, as shown in Figure~\ref{fig:figure2}. 
First, \textit{scene-grounded task initialization} (Section~\S\ref{sec:4.1}) transforms static observations into structured task repositories. Next, we detail the \textit{daytime exploration} (Section~\S\ref{sec:4.2}) and \textit{nighttime consolidation} (Section~\S\ref{sec:4.3}) phases, where the planner and simulator undergo joint online discovery and offline preference alignment. Finally, \textit{curriculum evolution} (Section~\S\ref{sec:4.4}) autonomously scales task complexity to ensure a stable learning trajectory.

\subsection{Scene-Grounding Task Initialization}
\label{sec:4.1}

To initiate the evolutionary loop from unlabeled images, \ourmethod first transforms raw images into structured, actionable task repositories, ensuring that exploration is grounded within the physical affordances of the observed scene.

\vspace{-0.4em}
\paragraph{Structured Scene Parsing.}
Given a seed image $I$, the planner $\mathcal{P}$ extracts a structured scene representation $S(I)$. This representation encapsulates essential entities and their spatial configurations, including: \ding{182} \textbf{objects} $\{o_k\}$ identified in the scene; \ding{183} \textbf{spatial relations} (\textit{e.g.}, on, in, near) that define the environmental topology; and \ding{184} \textbf{affordance priors} (\textit{e.g.}, pickable, openable) that constrain the action space.
To ensure robustness against perceptual errors or hallucinations in $\mathcal{P}$, a \textit{self-consistency voting} mechanism \citep{wang2023selfconsistency} is implemented, which has been widely proven effective in previous works \citep{guo2025deepseek, li-etal-2025-revisiting-self, hong2025slim, wan2025reasoning}. Specifically, $m=8$ independent parsing samples $\{S_j(I)\}_{j=1}^m$ are drawn, with only majority-consistent entities and relations retained to ensure a reliable foundation.

\vspace{-0.4em}
\paragraph{Task Template Instantiation.}
Following the widely adopted BridgeData V2 \citep{ebert2021bridge, zhang2025mind, fu2025learning} taxonomy, $S(I)$ is mapped into $13$ fundamental task templates $\mathcal{T}$ (\textit{e.g.}, {\small\texttt{"pick-and-place"}}, {\small\texttt{"stacking"}}), which serve as the building blocks for task initialization. $\mathcal{P}$ then instantiates and composes these primitives into structured plans. For instance, identified spatial and affordance relations may yield a composite task: {\small\texttt{"pick(bowl)} $\rightarrow$ \texttt{place(bowl, \text{rel}=on(table))} $\rightarrow$ \texttt{push(spoon, \text{rel}=in(cabinet))"}}. This hierarchical instantiation not only ensures task feasibility but also enables the generation of high-difficulty tasks through the composition of multiple basic actions.

\vspace{-0.4em}
\paragraph{Atomic-Action Difficulty Scoring.}
To facilitate difficulty-based curriculum evolution, each instantiated plan $\pi = \langle a_1, \dots, a_n \rangle$ is decomposed into a sequence of atomic actions $a_i \in \mathcal{A}$ (\textit{e.g.}, {\small\texttt{"grasp"}}, {\small\texttt{"lift"}}), each corresponding to a specific motion segment in the subsequent video generation. The difficulty of a task is quantified by $D(\pi|I)=\sum_{a_i \in \pi} c(a_i)$, the cumulative cost of its constituent actions. These scores provide a structured basis for binning tasks into difficulty levels, enabling the progressive evolution strategy in \S\ref{sec:4.4}. 

\subsection{Daytime Learning: Online Exploration}
\label{sec:4.2}

In the daytime phase, \ourmethod performs staged online exploration to jointly evolve the simulator $\mathcal{S}$ and the planner $\mathcal{P}$. 
By iteratively interleaving the daytime learning of $\mathcal{S}$ and $\mathcal{P}$, \ourmethod aligns the planner's high-level reasoning with the simulator's physical execution capabilities.

\vspace{-0.4em}
\paragraph{Simulator Daytime Training.}
The first stage focuses on improving the physical fidelity of the simulator $\mathcal{S}$, which serves as the foundation for verifying task execution. For each task $\tau$ initialized in Section~\S\ref{sec:4.1}, we sample $K$ video trajectories $\{V_1, \dots, V_K\}$ from $\mathcal{S}$, conditioned on the same task prompt $G \leftarrow \tau$. These trajectories are evaluated to identify relative advantages within the group using GRPO, which optimizes $\mathcal{S}$ by maximizing:
\begin{equation}
    \mathcal{J}_\text{Daytime}(\mathcal{S}) = \mathbb{E}_{\tau \sim \mathcal{D}, \{V_k\} \sim \mathcal{S}} \left[ \frac{1}{K} \sum_{k=1}^K \text{clip} \left( \frac{\mathcal{S}(V_k|\pi)}{\mathcal{S}_\text{old}(V_k|\pi)}, 1-\epsilon, 1+\epsilon \right) \hat{A}_k \right],
\end{equation}
where the advantage $\hat{A}_k$ is computed as $\hat{A}_k = R(V_k) - \frac{1}{K}\sum_{j=1}^K R(V_j)$. Here, $R(V)$ is a reward signal provided by $\mathcal{P}$ that evaluates the quality of $V$ based on semantic and physical alignment with the task, which is detailed in the following section. By iteratively refining $\mathcal{S}$, this stage improves its ability to generate physically consistent trajectories for tasks at a base difficulty level $D$.

\vspace{-0.4em}
\paragraph{Planner Daytime Training.}
While $\mathcal{S}$ is anchored by physical grounding, the planner $\mathcal{P}$ leverages VLM's abstract reasoning to transcend immediate physical feasibility. This capacity enables $\mathcal{P}$ to explore tasks beyond $\mathcal{S}$'s current limits. To this end, once $\mathcal{S}$ stabilizes at difficulty $D$, \ourmethod evolves $\mathcal{P}$ to handle more long-horizon tasks $\mathcal{T}_{\text{high}}$ (\textit{e.g.}, making a burger) with complexity $(D, 2D]$.

For each task $\tau \in \mathcal{T}_{\text{high}}$, $\mathcal{P}$ generates multiple candidate plans $\{\pi_1, \dots, \pi_K\}$, each decomposed into a sequence of atomic actions $\langle a_1, \dots, a_n \rangle$.
To reduce computational costs and mitigate potential hallucinations caused by over-reliance on $\mathcal{S}$, we propose a \textbf{selective simulation strategy}. Specifically, the self-consistency voting mechanism selects the most consistent plan $\pi^*$ for validation. $\pi^*$ is then executed in $\mathcal{S}$ via \textit{segment-wise simulation}, where each segment is constrained to difficulty $\leq D$ to stay within $\mathcal{S}$'s current capability.
The planner $\mathcal{P}$ is also optimized using GRPO with the reward:
\begin{equation}
    \hat{R}(\pi, \tau) = 1[\pi = \pi^*]\cdot (1+\eta \cdot R(\mathcal{S}(\pi^*))),
\end{equation}
where $1[\pi = \pi^*]$ filters for the consensus plan, and $R(\mathcal{S}(\pi^*))$ serves as a \textit{reward shaping} term providing physical feedback. This mechanism prevents $\mathcal{P}$ from adopting executionally infeasible logic while insulating it from potential simulator hallucinations via the multiplicative binary gate. By grounding abstract reasoning in physical constraints, this stage ensures that $\mathcal{P}$'s abstract reasoning remains aligned with the physical boundaries established by $\mathcal{S}$.

\vspace{-0.4em}
\paragraph{Semantic-controlled Multi-Granular Reward.}
To better supervise the evolution of $\mathcal{S}$ and $\mathcal{P}$, \ourmethod introduces a \textit{semantic-controlled multi-granular reward} $R(\cdot)$. Unlike relying solely on coarse visual-language alignment, our reward explicitly prioritizes \textit{semantic faithfulness} as a modulator mechanism to better capture subtle manipulation failures:
\begin{equation}
    R(V) = \mathbb{I}_\text{sem} \cdot \left( s_F + w_s s_S + s_E \right),
\end{equation}
where $\mathbb{I}_{\text{sem}}$ is a \textbf{semantic-alignment indicator}.
Specifically, instead of drafting a new prompt from scratch, $\mathcal{P}$ acts as a critic that inspects the trajectory $V$ and selectively modifies only the conflicting parts of the original goal $G$ to produce a revised prompt $G'$. The similarity $\text{Sim}(G, G')$ then serves as $\mathbb{I}_{\text{sem}}$, ensuring physical scores are proportionally suppressed upon semantic deviation.

To ensure numerical stability and preclude reward-hacking, the physical reward components are formulated as \textit{binary} signals:

\vspace{-0.8em}
\begin{itemize}[leftmargin=1.6em]
    \item[\ding{111}] \textbf{Frame-level Consistency} \textbf{(}$s_F$\textbf{)} penalizes discontinuities. $s_F=1$ only if object persistence and spatial smoothness are maintained across all frames.
    \item[\ding{111}] \textbf{Segment-level Execution} \textbf{(}$s_S$\textbf{)} is computed as $\frac{1}{M}\sum_{i=1}^M 1[a_i \in Seg_i]$, where $w_s = 1/M$ normalizes the action-wise binary detection.
    \item[\ding{111}] \textbf{Episode-level Success} \textbf{(}$s_E$\textbf{)} indicates final task achievement.
\end{itemize}
\vspace{-0.8em}

\noindent These discretized constraints provide robust supervision during daytime exploration and establish clear failure criteria for subsequent nighttime learning.

\begingroup
\setlength{\tabcolsep}{2.6pt}
\begin{table*}[t]
\vspace{-1.4em}
\renewcommand{\arraystretch}{1.16}
  \centering
  \caption{Evaluation of \ourmethod's Simulator $\mathcal{S}$ on BridgeData V2 \citep{ebert2021bridge} test set. \raisebox{-0.3ex}{\includegraphics[width=0.02\textwidth]{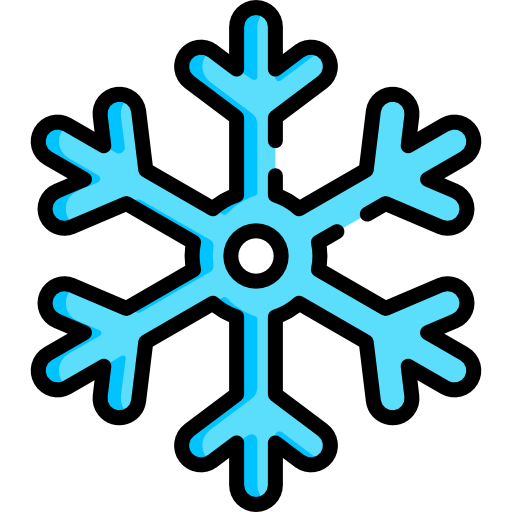}} indicates a frozen Planner $\mathcal{P}$. The best results are highlighted in \textbf{bold}.}
  \centering
  \vspace{-0.7em}
  \scriptsize
   \begin{tabular}{l ccc ccc ccc}
     \hlineB{2.5}
     \multirow{2}{*}{\textbf{Method}} & \multicolumn{3}{c}{\textbf{Level-1}} & \multicolumn{3}{c}{\textbf{Level-2}} & \multicolumn{3}{c}{\textbf{Level-3}}\\
     \cmidrule(lr){2-4} \cmidrule(lr){5-7} \cmidrule(lr){8-10}
     & \textbf{VBench$\uparrow$} & \textbf{Success$\uparrow$} & \textbf{User Pref.$\uparrow$} & \textbf{VBench$\uparrow$} & \textbf{Success$\uparrow$} & \textbf{User Pref.$\uparrow$} & \textbf{VBench$\uparrow$} & \textbf{Success$\uparrow$} & \textbf{User Pref.$\uparrow$} \\
     \hlineB{1.5}
     RoboDreamer  & 0.837 & 0.435 & 0.078 & 0.792 & 0.362 & 0.048 & 0.776 & 0.302 & 0.026 \\
     Wow-1-DiT  & 0.844 & 0.467 & 0.094 & 0.806 & 0.390 & 0.070 & 0.789 & 0.326 & 0.048 \\
     DreamDojo  & 0.848 & 0.500 & 0.136 & 0.824 & 0.416 & 0.098 & 0.813 & 0.347 & 0.094 \\
     Wow-1-Wan  & 0.846 & 0.519 & 0.142 & 0.820 & 0.439 & 0.106 & 0.806 & 0.364 & 0.076 \\
     \hline
     Wan2.2-TI2V  & 0.844 & 0.477 & 0.076 & 0.798 & 0.395 & 0.056 & 0.786 & 0.324 & 0.026 \\
     $+$ SFT (Cold Start) & 0.846 & 0.491 & 0.106 & 0.820 & 0.409 & 0.096 & 0.802 & 0.341 & 0.056 \\
     $+$ \raisebox{-0.3ex}{\includegraphics[width=0.02\textwidth]{figure/snowflake.png}} Planner $\mathcal{P}$ & 0.850 & 0.640 & 0.170 & 0.839 & 0.561 & 0.238 & 0.825 & 0.483 & 0.314 \\
     \rowcolor{cyan!10}
     $+$ \textbf{\ourmethod (Ours)} & \textbf{0.852} & \textbf{0.668} & \textbf{0.198} & \textbf{0.841} & \textbf{0.591} & \textbf{0.288} & \textbf{0.828} & \textbf{0.505} & \textbf{0.360}  \\
     \rowcolor{cyan!10}
     $\Delta\%/\Delta\%/\Delta$ & 0.9 & 40.0 & 0.122 & 5.1 & 49.6 & 0.232 & 5.3 & 55.9 & 0.334 \\
     \hlineB{2.5}
   \end{tabular}
  \label{tab:table1}
  \vspace{-1em}
\end{table*} 
\endgroup

\begingroup
\setlength{\tabcolsep}{1pt}
\begin{table*}[t]
\renewcommand{\arraystretch}{1.16}
  \centering
  \caption{Evaluation on \ourmethod's Planner $\mathcal{P}$ on EB-ALFRED and EB-Habitat \citep{yang2025embodiedbench}.}
  \centering
  \vspace{-0.7em}
  \scriptsize
   \begin{tabular}{l ccccccc ccccccc}
     \hlineB{2.5}
     \multirow{2}{*}{\textbf{Method}} & \multicolumn{7}{c}{\textbf{EB-ALFRED}} & \multicolumn{7}{c}{\textbf{EB-Habitat}}\\
     \cmidrule(lr){2-8} \cmidrule(lr){9-15}
     & \textbf{Base$\uparrow$} & \textbf{Comm.$\uparrow$} & \textbf{Comp.$\uparrow$} & \textbf{Vis.$\uparrow$} & \textbf{Spatial$\uparrow$} & \textbf{Long$\uparrow$} & \textbf{Avg.$\uparrow$} & \textbf{Base$\uparrow$} & \textbf{Comm.$\uparrow$} & \textbf{Comp.$\uparrow$} & \textbf{Vis.$\uparrow$} & \textbf{Spatial$\uparrow$} & \textbf{Long$\uparrow$} & \textbf{Avg.$\uparrow$}\\
     \hlineB{1.5}
     REBP & 54 & 42 & 46 & 28 & 38 & 6 & 35.6 & 50 & 6 & 18 & 14 & 14 & 8 & 18.3 \\
     RoboGPT-R1  & 62 & 56 & 64 & 50 & 50 & 50 & 55.3 & 64 & 8 & 18 & 20 & 12 & 10 & 22.0 \\
     WAP  & 66 & 62 & 70 & 56 & 52 & 70 & 62.7 & - & - & - & - & - & - & - \\
     RoboAgent  & 72 & 48 & 64 & 78 & 60 & 80 & 67.0 & - & - & - & - & - & - & 22.3 \\
     \hline
     Qwen3-VL  & 28 & 20 & 26 & 32 & 20 & 26 & 25.3 & 68 & 16 & 38 & 10 & 26 & 20 & 29.7 \\
     $+$ \raisebox{-0.3ex}{\includegraphics[width=0.02\textwidth]{figure/snowflake.png}} Simulator $\mathcal{S}$ & 56 & 46 & 64 & 56 & 46 & 66 & 55.7 & 76 & 26 & 58 & 44 & 42 & 48 & 49.0 \\
     \rowcolor{cyan!10}
     $+$ \textbf{\ourmethod (Ours)} & \textbf{60} & \textbf{52} & \textbf{70} & \textbf{62} & \textbf{52} & \textbf{74} & \textbf{61.7} & \textbf{82} & \textbf{28} & \textbf{64} & \textbf{48} & \textbf{46} & \textbf{58} & \textbf{54.3}  \\
     \rowcolor{cyan!10}
     $\Delta$ & 32 & 32 & 44 & 30 & 32 & 48 & 36.4 & 14 & 12 & 26 & 38 & 20 & 38 & 24.6 \\
     \hlineB{2.5}
   \end{tabular}
  \label{tab:table2}
\end{table*} 
\endgroup

\subsection{Nighttime Learning: Offline Consolidation}
\label{sec:4.3}

While daytime exploration provides behavioral breadth, the gathered trajectories often contain physical hallucinations and planning fallacies. Nighttime learning serves as a cortical consolidation phase, transforming these raw experiences into high-value preference pairs to refine both $\mathcal{S}$ and $\mathcal{P}$.

\vspace{-0.4em}
\paragraph{Simulator Nighttime Training.}
The primary objective for $\mathcal{S}$ during nighttime is to suppress intrinsic physical hallucinations by learning from "near-miss" failures. Using the binary multi-granular rewards from Section~\S\ref{sec:4.2}, we construct video preference pairs $(V^+, V^-)$: \ding{182} \textbf{positive samples (}$V^+$\textbf{)}: trajectories with high cumulative rewards ($s_E=1$ and $\mathbb{I}_\text{sem}=1$), representing physically consistent and task-aligned executions; and \ding{183} \textbf{negative samples (}$V^-$\textbf{)}: critically, we select hard negatives that satisfy at least one validity criterion (\textit{e.g.}, $s_F=1$ or $w_s s_S=1$) but fail the overall task ($s_E=0$). By contrasting these pairs, $\mathcal{S}$ learns to distinguish visually plausible but invalid motions from physically sound ones. The optimization follows the DPO objective:
\begin{equation}
    \mathcal{L}_\text{Nighttime}(\mathcal{S}) = -\mathbb{E}_{(V^+, V^-)} \left[ \log \sigma \left( \beta \log \frac{\mathcal{S}(V^+|\pi)}{\mathcal{S}_\text{ref}(V^+|\pi)} - \beta \log \frac{\mathcal{S}(V^-|\pi)}{\mathcal{S}_\text{ref}(V^-|\pi)} \right) \right],
\end{equation}
where $\mathcal{S}_{\text{ref}}$ is the simulator from the previous iteration. This ensures that $\mathcal{S}$ progressively aligns with the manifold of physical reality.

\begin{figure*}[!t]
\centering
\vspace{-1.6em}
\includegraphics[width=\linewidth]{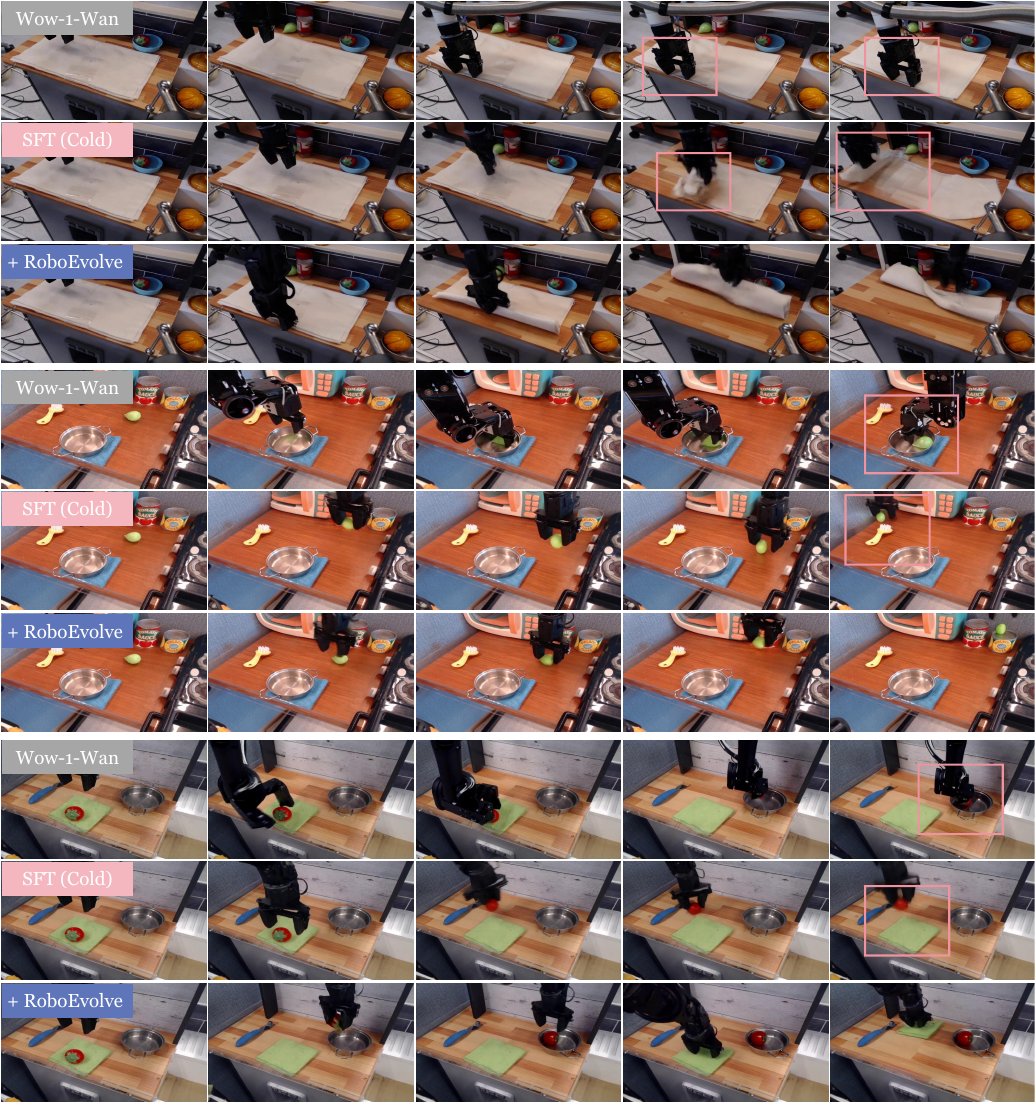}
\vspace{-1.4em}
\caption{Qualitative results of \ourmethod. Baseline indicates the SFT cold started one. {\small(\textbf{\textit{Top}}) \textit{\textbf{Level-1}: The robotic arm folds the cloth in half.} (\textbf{\textit{Middle}}) \textit{\textbf{Level-2}: The robotic arm picks up the green ball, and places it on the can to the right rear.} (\textbf{\textit{Bottom}}) \textit{\textbf{Level-3}: The robotic arm picks up the tomato, places the tomato into the pot, and finally moves the green cloth to the left of the pot.} } }
\label{fig:figure_qualitative}
\vspace{-1.4em}
\end{figure*}

\vspace{-0.4em}
\paragraph{Planner Nighttime Training.}
To maximize the utility of the sparse interaction data collected during daytime, we propose a \textbf{hierarchical preference optimization} strategy, which enables $\mathcal{P}$ to extract rich preference signals across three distinct cognitive dimensions:

\vspace{-0.6em}
\begin{itemize}[leftmargin=1.6em]
    \item[\ding{182}] \textbf{Planning-level (}$\mathcal{D}_P$\textbf{)}: Given an initial image $I$, $\mathcal{P}$ learns to prioritize logical consistency, which learns to prefer the consensus plan $\pi^*$ from majority voting over suboptimal runner-up candidates.
    \item[\ding{183}] \textbf{Understanding-level (}$\mathcal{D}_U$\textbf{)}: Given a high-reward video $V^+$, $\mathcal{P}$ enhances its visual grounding, which favor the validated goal $G$ over erroneous back-translations, rectifying perceptual misconceptions.
    \item[\ding{184}] \textbf{Transition-level (}$\mathcal{D}_T$\textbf{)}: Given a state pair $(f_1, f_T)$ extracted from $V^+$, $\mathcal{P}$ internalizes causality by predicting the underlying intent, where $\mathcal{P}$ should correctly infer the plan $\pi^*$ over misidentified intents to internalize causality.
\end{itemize}
\vspace{-0.6em}

\noindent The planner is refined by minimizing the cumulative objective $\mathcal{L}_{\text{Nighttime}}(\mathcal{P})$:
\begin{equation} 
-\sum_{k \in \{P, U, T\}} \mathbb{E}_{(\mathbf{c}, \pi^+, \pi^-) \sim \mathcal{D}_k} \left[ \log \sigma \left( \beta \log \frac{\mathcal{P}(\pi^+ | \mathbf{c})}{\mathcal{P}_{\text{ref}}(\pi^+ | \mathbf{c})} - \beta \log \frac{\mathcal{P}(\pi^- | \mathbf{c})}{\mathcal{P}_{\text{ref}}(\pi^- | \mathbf{c})} \right) \right].
\end{equation}
By jointly refining $\mathcal{S}$ and $\mathcal{P}$, nighttime learning converts unsuccessful daytime attempts into valuable supervisory signals, closing the loop between imagination and reality and preparing the system for the next cycle of exploration.

\subsection{Dual-Phase Curriculum Evolution}
\label{sec:4.4}
To ensure a stable and progressive transition from simple manipulations to complex behaviors, \ourmethod incorporates a \textbf{progress-based curriculum} that autonomously scales task complexity. We discretize the difficulty space into $B$ bins based on the score $D(\pi|I)$. For each bin $b$, we track the success rate $S(b)$ and define the learning progress as $P_k(b) = S_k(b) - S_{k-\Delta}(b)$. To select the optimal difficulty for the next exploration phase, we employ an \textit{upper confidence bound} strategy \citep{auer2002using}:
\begin{equation}
b^*_k = \text{argmax}_{b} \left( P_k(b) + \lambda \sqrt{\frac{\log \sum_j n_k(j)}{n_k(b) + 1}} \right),
\label{eq:curriculum}
\end{equation}
where $n_k(b)$ is the sampling count and $\lambda$ weights the exploration. As success rates on simpler tasks saturate ($P_k(b) \to 0$), this mechanism inherently shifts the sampling budget toward higher-complexity frontiers, enabling continuous, manual-free capability scaling.

\section{Experiments}
\label{sec:experiment}
\vspace{-0.6em}

In this section, we conduct extensive experiments to answer the following key research questions: (\textbf{RQ1}) Does \ourmethod's co-evolutionary framework outperform existing static paradigms?
(\textbf{RQ2}) Can \ourmethod's progressive dual-phase curriculum enable robust continual learning?
(\textbf{RQ3}) Can \ourmethod achieve consistent performance gains by scaling the number of unlabeled seed images?
(\textbf{RQ4}) Can \ourmethod’s co-evolutionary framework improve low-level action execution when instantiated with an action policy?

\vspace{-0.6em}
\subsection{Experimental Settings}
\vspace{-0.6em}

\paragraph{Baselines.} We instantiate \ourmethod with Wan2.2-TI2V-5B \citep{wan2025wan} as the simulator $\mathcal{S}$ and Qwen3-VL-4B \citep{bai2023qwen} as the planner $\mathcal{P}$, sized specifically to meet the lightweight deployment demands of embodied agents \citep{grover2026embodied, takagi2026anolevla, kim2024openvla}. To rigorously isolate the efficacy of our co-evolutionary paradigm, we evaluate against key systemic variants: (\textbf{\textit{i}}) vanilla base models, (\textbf{\textit{ii}}) daytime-only, (\textbf{\textit{iii}}) nighttime-only, and (\textbf{\textit{iv}}) decoupled single-agent (simulator/planner-only) baselines. Moreover, we incorporate domain-specific methods as external references \citep{zhou2024robodreamer, chi2025wow, gao2026dreamdojo, xu2026roboagent, liu2025robogpt, shi2026worldaware, wu2025reinforced}.

\vspace{-0.7em}
\paragraph{Evaluations.} We evaluate \ourmethod on established benchmarks tailored for both system components: \textbf{(I) Simulator}: We utilize the BridgeData V2 \citep{ebert2021bridge} test set, featuring diverse manipulation skills (\textit{e.g.}, \texttt{move}, \texttt{pick}). To rigorously assess scalability, we evaluate across three complexity tiers: single atomic skills (Level-1) and compositionally chained complex tasks (Levels 2-3). Following MIND-V \citep{zhang2025mind}, our metrics include VBench \citep{huang2024vbench} for general visual quality (averaged from Aesthetic Quality, Imaging Quality, Temporal Flickering, Motion Smoothness, Subject Consistency, and Background Consistency), Gemini-2.5-Pro \citep{comanici2025gemini} for Task Success Rate, and User Preference for detailed embodied alignment. \textbf{(II) Planner}: We assess reasoning capabilities on EB-ALFRED and EB-Habitat \citep{yang2025embodiedbench}, two standard embodied planning benchmarks for multi-step household tasks.

\vspace{-0.7em}
\paragraph{Implementation Details.} To facilitate stable exploration, the simulator $\mathcal{S}$ is initialized with a SFT cold start on the BridgeData V2 \citep{ebert2021bridge} training split. The optimization pipelines for $\mathcal{S}$ are built upon Flow-Factory \citep{ping2026flow} and Flow-DPO \citep{liu2026improving}, while the planner $\mathcal{P}$ is trained using the TRL library \citep{vonwerra2020trl}. During daytime GRPO, the rollout size is set to $K=16$ and the reward shaping coefficient is set to $\eta=0.2$. In our core evaluation setting, the dual-phase curriculum scales tasks up to a maximum difficulty of $D=3$ ($\lambda=0.1$). Bootstrapping with $500$ unlabeled seed images yields $877$ $D=1$ atomic tasks, which combine to $3,363$ ($D=2$) and $9,228$ ($D=3$) composite tasks. All experiments are conducted on NVIDIA A800 GPUs.

The settings above describe the VLM–VGM instantiation; the setup for the action-policy extension to low-level control is detailed in Section~\S\ref{sec:libero_extension}.

\begin{figure*}[!t]
\centering
\vspace{-1em}
\includegraphics[width=\linewidth]{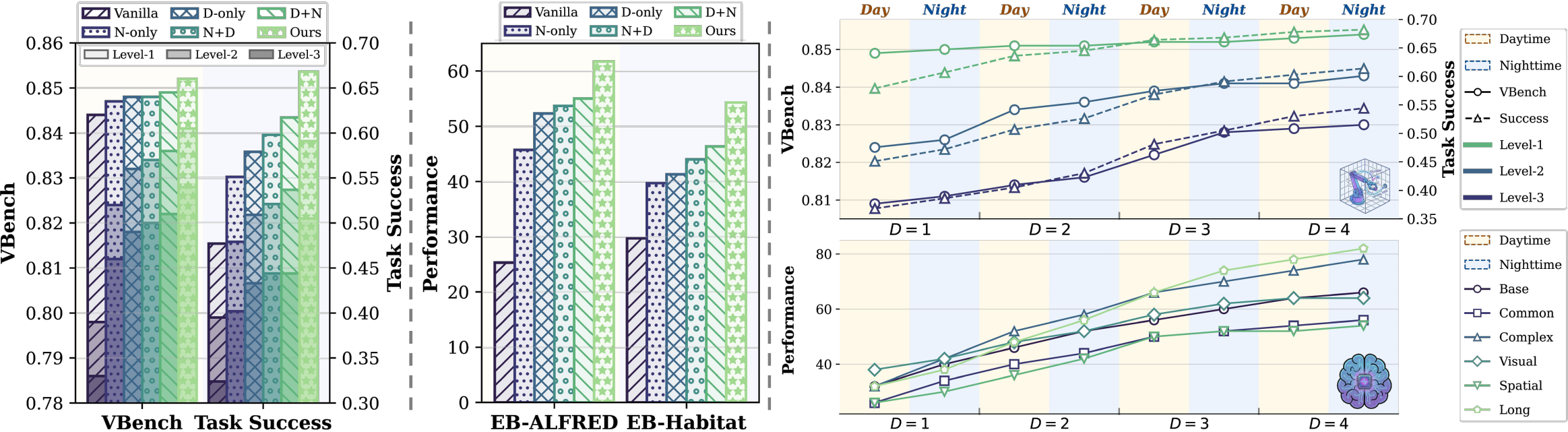}
\vspace{-1.4em}
\caption{(\textit{\textbf{Left \& Middle}}) Dual-phase evolution loop ablations for Simulator $\mathcal{S}$ and Planner $\mathcal{P}$. Sequential baselines (\textit{e.g.}, "D+N" denotes completing all Daytime exploration prior to Nighttime consolidation) underperform, highlighting the necessity of our interleaved evolution. (\textit{\textbf{Right}}) Curriculum evolution curves of \ourmethod. While our core evaluation caps at difficulty $D=3$, we extend the curriculum to $D=4$ here to demonstrate \ourmethod's robust continual learning capabilities.}
\label{fig:figure_main}
\vspace{-0.3em}
\end{figure*}

\subsection{Main Results}

To answer \textbf{RQ1}, we first conduct a comprehensive evaluation on BridgeData V2 for Simulator $\mathcal{S}$ (Table~\ref{tab:table1}), and on EB-ALFRED and EB-Habitat for Planner $\mathcal{P}$ (Table~\ref{tab:table2}), complemented by qualitative visualizations in Figures~\ref{fig:figure_qualitative},~\ref{app_fig:app_quali1},~\ref{app_fig:app_quali2} and~\ref{app_fig:app_quali3}. Key observations are summarized as follows:

\textbf{Obs.}\ding{182} \textbf{\ourmethod significantly outperforms static paradigms, with performance gains amplifying on complex tasks.}
As shown in Table~\ref{tab:table1}, static baselines and decoupled variants (\textit{e.g.}, w/ a frozen $\mathcal{P}$) struggle with physical grounding as complexity increases, whereas \ourmethod's mutually reinforcing loop yields substantial improvements across all metrics. Crucially, its relative Task Success gains over the static base model amplify dramatically on harder tasks, scaling from $40.0\%$ (Level-1) to $49.6\%$ (Level-2), and peaking at $55.9\%$ (Level-3). A parallel trend emerges in $\mathcal{P}$'s reasoning capabilities, as shown in Table~\ref{tab:table2}, where \ourmethod elevates the generalist Qwen3-VL base model by an average of $36.4$ and $24.6$ absolute points on EB-ALFRED and EB-Habitat without relying on domain-specific training. It significantly outperforms the static variants, particularly in spatial and long-horizon reasoning, securing highly competitive results against in-domain experts.
Qualitatively, Figure~\ref{fig:figure_qualitative} reveals that the static SFT baseline struggles with severe object distortions and often halts prematurely during multi-stage instructions. Furthermore, Figures~\ref{app_fig:app_quali1},~\ref{app_fig:app_quali2} and~\ref{app_fig:app_quali3} show that while the \textit{Daytime-Only} variant manages to attempt more complex sequences, it remains plagued by catastrophic physical hallucinations, such as sudden object disappearance. In stark contrast, \ourmethod robustly preserves structural integrity and semantic alignment, ensuring continuous physical plausibility across complex executions. Additional evaluation of the video simulator on VideoPhy-1 \citep{bansal2025videophy} and VideoPhy-2 \citep{bansal2026videophy} is provided in Appendix~\S\ref{app:videophy}, where \ourmethod improves both semantic adherence and physical commonsense scores.

\subsection{Curriculum Scaling Analysis}

To answer \textbf{RQ2}, we further dissect the evolutionary dynamics of \ourmethod, focusing on the interplay between phase scheduling in Figure~\ref{fig:figure_main} (\textit{Left \& Middle}) and task difficulty scaling in Figure~\ref{fig:figure_main} (\textit{Right}). Our findings are summarized as follows:

\textbf{Obs.}\ding{183} \textbf{Nighttime consolidation is a critical policy optimization stabilizer.} To isolate the mechanistic advantage of our dual-phase evolution loop, we evaluate structural ablations for both $\mathcal{S}$ and $\mathcal{P}$ in Figure~\ref{fig:figure_main} (\textit{Left \& Middle}). Baselines relying on isolated phases (\textit{e.g.}, Daytime-only) quickly saturate, as the system accumulates uncorrected physical hallucinations during unchecked exploration. Furthermore, the sequential baseline ("D+N"), which delays nighttime consolidation until all daytime exploration concludes, suffers from irreversible policy degradation, yielding significantly lower Task Success compared to our strategy. In contrast, \ourmethod's tightly interleaved sleep-wake cycles provide timely rectification. By aggressively penalizing out-of-distribution behaviors through "near-miss" failures during nighttime failure learning, the consolidation phase acts as an indispensable stabilizer, preventing the model from collapsing during continuous exploration.

\textbf{Obs.}\ding{184} \textbf{\ourmethod demonstrates strong continual learning capability.} While interleaved phases ensure stability, our progressive curriculum drives scalable capability acquisition. Figure~\ref{fig:figure_main} (\textit{Right}) tracks the evolutionary trajectories across difficulty stages $D=1$ through $D=4$. Rather than collapsing under the severe reward sparsity typically encountered when directly tackling complex tasks, \ourmethod organically paces its learning. By autonomously advancing the curriculum only when foundational skills saturate, the framework maintains monotonic performance gains across all metrics for both $\mathcal{S}$ and $\mathcal{P}$. Notably, by extending the curriculum to $D=4$, we observe that the system still seamlessly masters increasingly complex, composite tasks without catastrophic forgetting of simpler atomic actions, highlighting its open-ended potential and robust continual learning capabilities.

\subsection{Data Efficiency Analysis}

\begin{wrapfigure}{r}{0.3\textwidth}
\vspace{-1.2em}
 \centering
 \includegraphics[width=\linewidth]{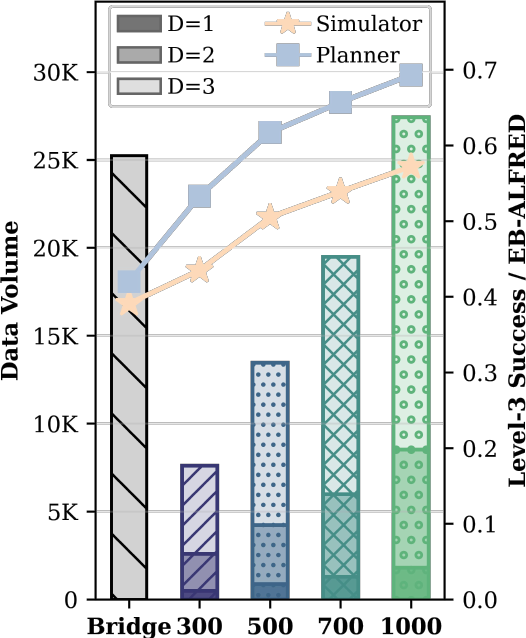}
  \vspace{-1.64em}
  \caption{Data scaling \textit{vs.} performance of \ourmethod. The Planner's score is scaled by $0.01$ for visual alignment.}
  \vspace{-3em}
  \label{fig:data_scaling}
\end{wrapfigure}

To answer \textbf{RQ3}, we further investigate the scaling behavior of \ourmethod by varying the number of initial unlabeled seed images (from $300$ to $1000$), as shown in Figure~\ref{fig:data_scaling}. To establish a rigorous baseline for data efficiency, we further benchmark our autonomous scaling against training directly on the raw BridgeData V2 ($\sim$$25$K manually annotated trajectories). Our key observations are:

\textbf{Obs.}\ding{185} \textbf{\ourmethod scales effectively using purely unlabeled seed images.}
As illustrated in Figure~\ref{fig:data_scaling}, scaling unlabeled seed images consistently drives complex data instantiation ($D=1$ to $3$), translating into monotonic performance gains for both $\mathcal{S}$ and $\mathcal{P}$. Crucially, starting with merely $300$ seeds, \ourmethod synthesizes only $\sim$$7.6$K trajectories, \textit{i.e.}, less than a third of the raw BridgeData ($\sim$$25$K), yet entirely surpasses its $\mathcal{S}$'s Level-3 Task Success and $\mathcal{P}$'s EB-ALFRED performance. This compelling result highlights that \ourmethod not only scales robustly without human intervention but also synthesizes high-density, task-relevant supervision that is significantly more effective than massive, static flat data collection.

\vspace{-0.7em}
\paragraph{Additional Analyses.} Appendix~\S\ref{app_results} provides further analyses of \ourmethod, including RL training dynamics (Figure~\ref{app_fig:app_Rl}) and ablations of the multi-granular reward, selective simulation, hierarchical preference optimization, and curriculum hyperparameter. We additionally evaluate semantic adherence and physical commonsense on VideoPhy-1 and VideoPhy-2 (Appendix~\S\ref{app:videophy}) and examine the effect of the base planner (Appendix~\S\ref{app:base_planner}).

\subsection{Extension to Low-Level Action Execution}
\label{sec:libero_extension}

To answer \textbf{RQ4}, we further instantiate \ourmethod with LingBot-VA \citep{li2026causal}, a world action model integrating visual dynamics prediction and action inference, and evaluate low-level control on LIBERO \citep{liu2023libero}, as shown in Table~\ref{tab:libero}. The experimental settings and observations are detailed as follows:

\begingroup
\setlength{\tabcolsep}{13.8pt}
\begin{table*}[!h]
\vspace{-0.4em}
\renewcommand{\arraystretch}{1.2}
  \centering
  \caption{Task success rates on LIBERO. \ourmethod starts from the cold-started LingBot-VA checkpoint; each cycle comprises daytime exploration and nighttime consolidation.}
  \centering
  \vspace{-0.7em}
  \footnotesize
   \begin{tabular}{lc|cccccc}
     \hlineB{2.5}
     \rowcolor{CadetBlue!20} 
     \textbf{Method} & \textbf{Data} & \textbf{Spatial$\uparrow$} & \textbf{Object$\uparrow$} & \textbf{Goal$\uparrow$} & \textbf{Long$\uparrow$} & \textbf{Avg.$\uparrow$} \\
     \hlineB{1.5}
     $\pi_0$~\citep{black2024pi_0} &  & 96.8 & 98.8 & 95.8 & 85.2 & 94.1   \\
     \rowcolor{lightgray!20} 
     OpenVLA~\citep{kim2024openvla} &  & 84.7 & 88.4 & 79.2 & 53.7 & 76.5  \\
     \hline
     LingBot-VA~\citep{li2026causal} &  &  &  &  &  &    \\
     \rowcolor{lightgray!20} 
     + Post-trained (SFT) & 256,000 & \underline{98.5} & \textbf{99.6} & \underline{97.2} & \textbf{98.5} & \underline{98.5}   \\
     \hline
     LingBot-VA~\citep{li2026causal} &  &  &  &  &  &    \\
     \rowcolor{lightgray!20} 
     + Cold-started (SFT) & 4,000 & 78.9 & 88.4 & 78.6 & 69.0 & 78.7   \\
     + \ourmethod-P1 & 500 & 96.8 & 96.8 & 93.2 & 88.4 & 93.8   \\
     \rowcolor{lightgray!20} 
     + \ourmethod-P2 & - & 98.2 & 98.4 & 96.0 & 92.4 & 96.2   \\
     + \ourmethod-P3 & - & \textbf{98.8} & \textbf{99.6} & \textbf{97.9} & \underline{98.4} & \textbf{98.7}  \\
     \hlineB{2.5}
   \end{tabular}
  \label{tab:libero}
  \vspace{-0.4em}
\end{table*} 
\endgroup

\textbf{Experimental Settings.}\quad In the original VLM-VGM setting, we use VLM-based rewards to examine whether the planner and simulator can improve through mutual supervision without external feedback during self-evolution. The dual-phase learning design also accommodates environment feedback when an executable action interface is available. We therefore instantiate \ourmethod with LingBot-VA, retaining the VLM planner and the alternating exploration-consolidation loop while using feedback from action execution in LIBERO to guide learning. This instantiation evaluates the applicability of \ourmethod to low-level control in a physics-based environment. Self-evolution starts from a cold-started LingBot-VA checkpoint and $500$ unlabeled seed images. We report task success across four LIBERO suites, with P1-P3 denoting successive day-night cycles and post-trained LingBot-VA serving as a supervised reference.

\textbf{Obs.\ding{186} \ourmethod extends effective co-evolution to low-level control.}
As shown in Table~\ref{tab:libero}, starting from cold-started LingBot-VA and $500$ unlabeled seed images, \ourmethod improves average task success from $78.7\%$ to $93.8\%$, $96.2\%$, and $98.7\%$ across successive day-night cycles. The largest gain occurs on the Long suite, where success increases from $69.0\%$ to $98.4\%$, highlighting improved execution of multi-stage tasks. With a 20.0-point average gain, the final model reaches performance comparable to the extensively supervised post-trained reference ($98.7\%$ \textit{vs.} $98.5\%$). These results show that iterative exploration and consolidation can substantially improve an action-capable model after limited supervised initialization. Moreover, the gains extend the empirical support for our learning design beyond the original VLM-based reward setting: the same dual-phase framework can incorporate feedback from executed actions to guide continued improvement under simulated physical dynamics. Together with the VLM-VGM results, this demonstrates the applicability of \ourmethod across both internally supervised generative learning and environment-grounded action learning.

\section{Conclusion}

In this work, we introduced \ourmethod, a co-evolutionary framework that mutually refines a VLM planner and a VGM simulator for data-efficient robotic manipulation learning. Starting from unlabeled seed images after model initialization, its dual-phase loop interleaves daytime exploration with nighttime failure consolidation under a progressive curriculum. Experiments show consistent improvements in video simulation and embodied planning, sustained gains as task complexity increases, and strong data efficiency: self-evolution from merely $300$ seeds surpasses the supervised BridgeData baseline on Level-3 simulator success and EB-ALFRED planning. Extending the framework to low-level control with environment feedback further improves average LIBERO task success from $78.7\%$ to $98.7\%$, reaching performance comparable to the extensively supervised reference. Together, these findings support co-evolution through exploration and failure consolidation as a scalable approach to both internally supervised generative learning and environment-grounded action learning.




\bibliography{main}
\bibliographystyle{tmlr}

\newpage
\appendix

\section{More Details of Scene-Grounded Task Initialization}
\label{app_sec:scene_details}


\subsection{Taxonomy of Task Templates and Atomic Actions}

To bridge the semantic gap between high-level reasoning and low-level physical execution, we construct a structured task taxonomy inspired by the BridgeData V2 \citep{ebert2021bridge} dataset. This taxonomy decomposes diverse manipulation skills into $13$ fundamental task templates $\mathcal{T}$, which are grounded into a comprehensive atomic-action space $\mathcal{A}$.

To enable the autonomous, horizon-based curriculum evolution detailed in Section~\ref{sec:4.4}, each atomic action $a_i \in \mathcal{A}$ is treated as an indivisible unit with a base execution cost of $c(a_i) = 1$. Consequently, the overall task difficulty $D(\tau|I) = \sum_{a_i \in \tau} c(a_i)$ directly corresponds to the task's compositional horizon length. By chaining these templates, the Planner $\mathcal{P}$ systematically instantiates tasks of scaling complexity. For instance, a single atomic task like "Opening a drawer" yields $D=1$, while a composite two-stage task such as "Open the drawer and pick up the apple" yields $D = c(\texttt{open}) + c(\texttt{pick}) = 2$. Table~\ref{tab:taxonomy} here categorizes these atomic actions and details their physical execution requirements, serving as the foundational building blocks for all evaluated multi-level benchmarks.

\begingroup
\setlength{\tabcolsep}{26pt}
\begin{table*}[!h]
\vspace{-0.6em}
\renewcommand{\arraystretch}{1.2}
  \centering
\caption{Taxonomy of atomic actions $\mathcal{A}$, and their physical execution descriptions.}
\footnotesize
\vspace{-0.7em}
\label{tab:taxonomy}
\begin{tabular}{ll}
\hlineB{2.5}
\rowcolor{CadetBlue!20}
 \textbf{Atomic Action $a_i \in \mathcal{A}$} & \textbf{Physical Execution Illustration}  \\
\hlineB{1.5}
\texttt{pick(X, grasp\_hint)} & Grasp and lift object \texttt{X} (approach, align, close, lift). \\
\rowcolor{lightgray!20} 
\texttt{place(X, target)} & Place \texttt{X} at \texttt{target} (region/relation) and release stably.  \\
\texttt{push(X, dir, dist)} & Make physical contact and push \texttt{X} along a direction. \\
\rowcolor{lightgray!20} 
\texttt{stack\_on(X, Y)} & Place \texttt{X} precisely on top of \texttt{Y} and stabilize.  \\
\texttt{wipe(area, tool, strokes)} & Wipe a specified \texttt{area} with a \texttt{tool} (sustained contact).  \\
\rowcolor{lightgray!20} 
\texttt{sweep(objs, tool, region)} & Sweep scattered \texttt{objs} into a target \texttt{region} using a \texttt{tool}.  \\
\texttt{fold(X, pattern)} & Fold cloth \texttt{X} (grasp edge/corner, flip, and flatten).  \\
\rowcolor{lightgray!20} 
\texttt{zip(Z, dir, length)} & Pull zipper \texttt{Z} along a direction (\texttt{zip} or \texttt{unzip}).  \\
\texttt{open(C, part, method)} & Open a container/mechanism \texttt{C} (drawer, door, box flap).   \\
\rowcolor{lightgray!20} 
\texttt{close(C, part, method)} & Close a container/mechanism \texttt{C}.  \\
\texttt{turn\_knob(K, angle)} & Rotate knob \texttt{K} to a specific state or angle.  \\
\rowcolor{lightgray!20} 
\texttt{toggle\_switch(S, state)} & Flip switch \texttt{S} to a desired \texttt{state}.   \\
\texttt{turn\_lever(L, angle)} & Pull or rotate lever \texttt{L} to a specific state or angle.  \\
\hlineB{2.5}
\end{tabular}%
\vspace{-1em}
\end{table*}
\endgroup

\subsection{More Details of Self-Consistency Voting}

\begin{wrapfigure}{r}{0.2\textwidth}
\vspace{-1.4em}
 \centering
 \includegraphics[width=\linewidth]{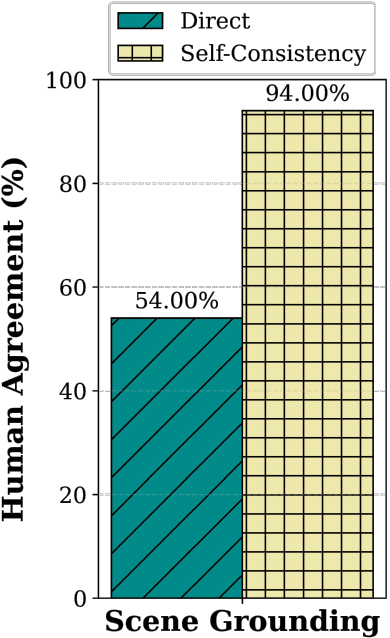}
  \vspace{-1.64em}
  \caption{Human evaluation of self-consistency scene-grounding.}
  \vspace{-2.2em}
  \label{app_fig:consistency}
\end{wrapfigure}

As introduced in Section~\ref{sec:4.1}, translating raw seed images into physically feasible task prompts requires precise and reliable scene understanding. Relying on a single-pass inference from VLMs often leads to object hallucinations or overlooked spatial constraints. 

To evaluate the efficacy of our self-consistency voting mechanism for scene grounding, we conduct a human verification study on $50$ randomly sampled seed images. Evaluators assessed the physical correctness of parsed objects, affordances, and spatial relations. As shown in Figure~\ref{app_fig:consistency}, a single direct VLM pass proves highly brittle, yielding merely a $54.00\%$ human agreement rate. In stark contrast, our self-consistency strategy drastically mitigates semantic hallucinations, elevating grounding accuracy to $94.00\%$ by retaining only majority-agreed entities.

\begin{figure*}[!h]
\centering
\vspace{-1em}
\includegraphics[width=\linewidth]{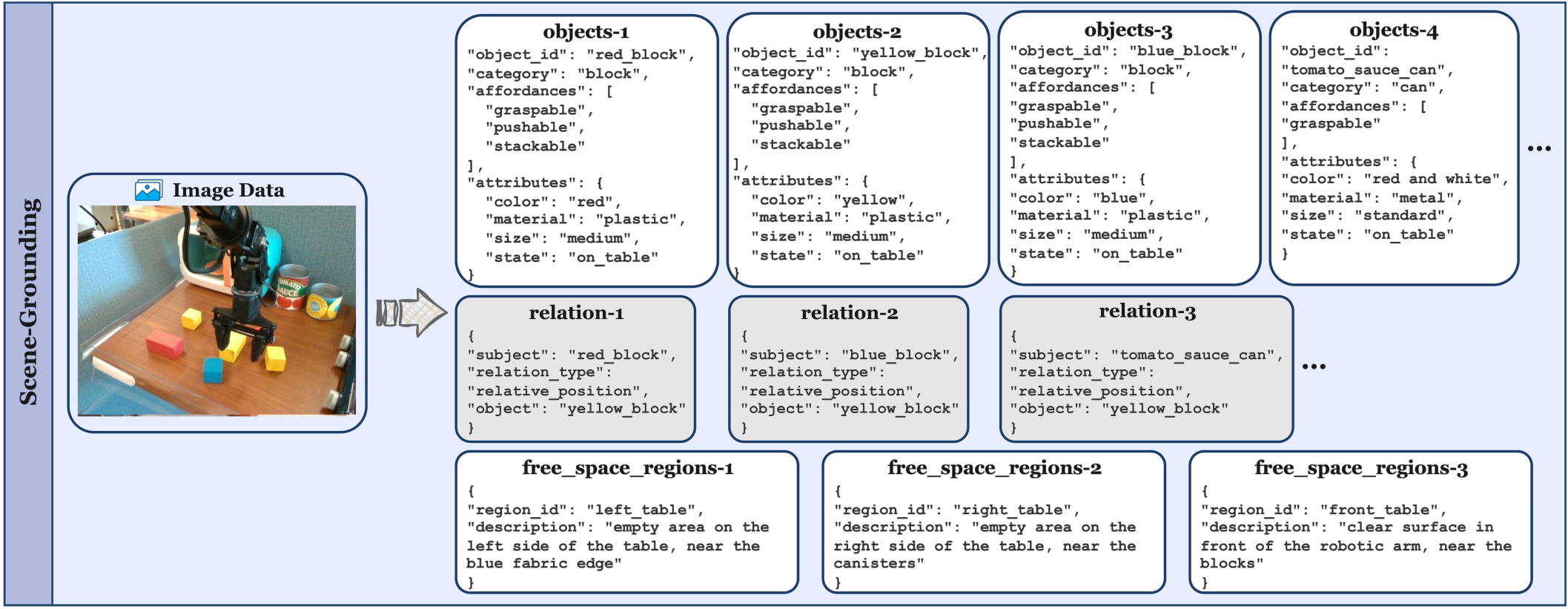}
\vspace{-1.6em}
\caption{Case demonstration of scene-grounding task initialization.}
\label{app_fig:app_b_2-1}
\vspace{-1.3em}
\end{figure*}

\begin{figure*}[!h]
\centering
\vspace{-1em}
\includegraphics[width=\linewidth]{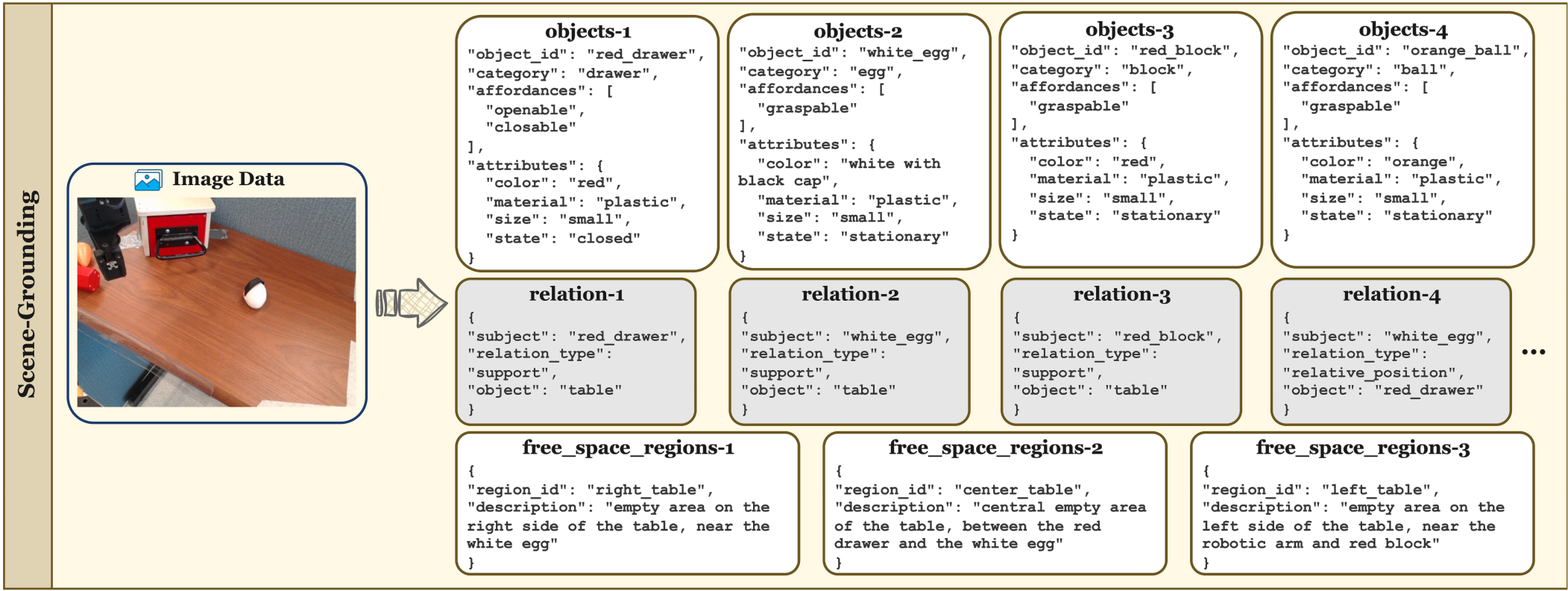}
\vspace{-1.6em}
\caption{Case demonstration of scene-grounding task initialization.}
\label{app_fig:app_b_2-2}
\vspace{-1.3em}
\end{figure*}

\begin{figure*}[!t]
\centering
\includegraphics[width=\linewidth]{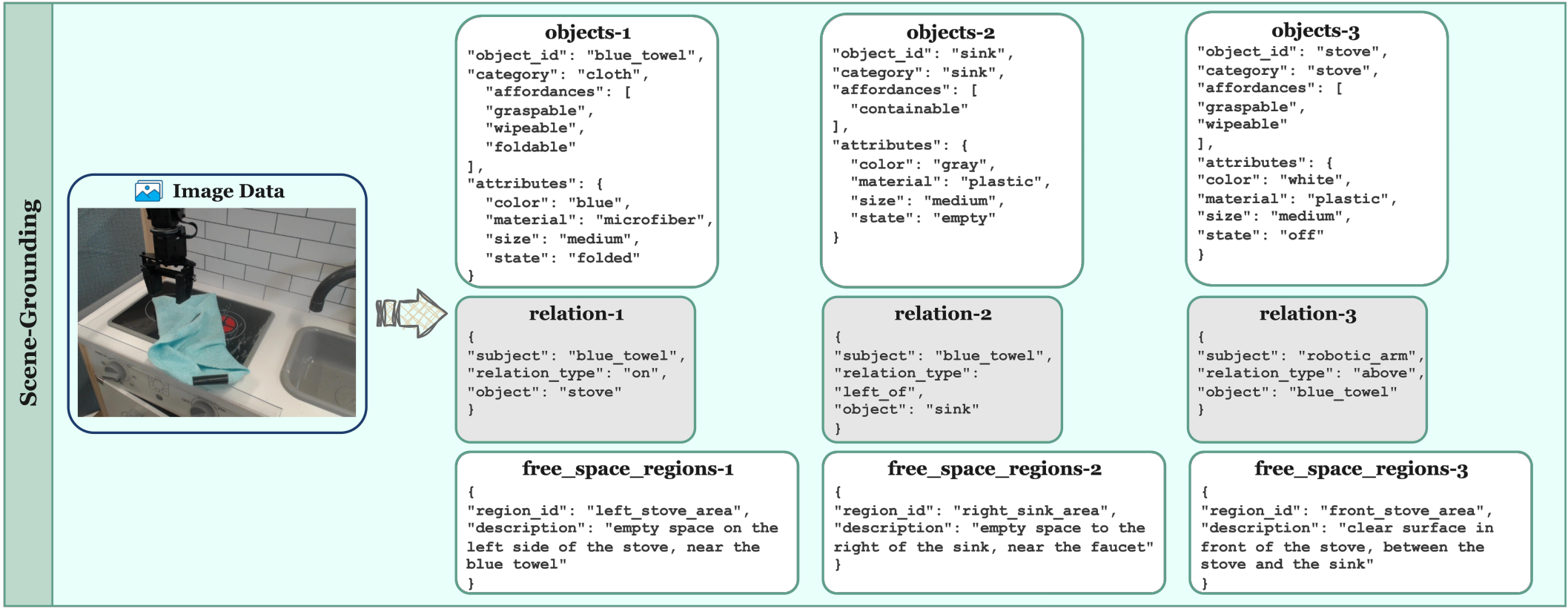}
\vspace{-1.6em}
\caption{Case demonstration of scene-grounding task initialization.}
\label{app_fig:app_b_2-3}
\vspace{-1.3em}
\end{figure*}

Furthermore, Figures~\ref{app_fig:app_b_2-1},~\ref{app_fig:app_b_2-2}, and~\ref{app_fig:app_b_2-3} qualitatively demonstrate this initialization process across diverse environments. The voting mechanism robustly decomposes raw visual inputs into structured symbolic representations, accurately capturing object attributes (\textit{e.g.}, \texttt{graspable}, \texttt{openable}), pair-wise relations, and actionable free-space regions. This precise, multi-dimensional scene grounding serves as the indispensable bedrock for synthesizing physically viable trajectories in the subsequent curriculum.

\section{Experimental Details of \ourmethod}
\label{app_sec:experimental_details}

\subsection{More Details of Semantic-Controlled Multi-Granular Reward}

As introduced in Section~\S\ref{sec:4.2}, the semantic-controlled multi-granular reward $R(\cdot)$ is pivotal for mitigating physical hallucinations and providing high-density supervisory signals during the daytime exploration phase. Relying on a single, holistic prompt to evaluate complex video trajectories often causes the vision-language Planner $\mathcal{P}$ to overlook subtle physical discontinuities or semantic deviations. To ensure robust and precise feedback, we systematically decouple the evaluation protocol into four distinct cognitive dimensions: \textbf{semantic-alignment} ($\mathbb{I}_{sem}$), \textbf{frame-level consistency} ($s_F$), \textbf{segment-level execution} ($s_S$), and \textbf{episode-level success} ($s_E$). Below, we detail the exact system prompts for evaluating these four granularities.

\begin{tcolorbox}[notitle, sharp corners, breakable, colframe=teal!40, colback=gray!4, 
       boxrule=3pt, boxsep=0.5pt, enhanced, 
       shadow={3pt}{-3pt}{0pt}{opacity=1,gray!10},
       title={Prompt for Semantic-alignment Indicator ($\mathbb{I}_\text{sem}$) Rewarding}]\label{box:prompt}
       \footnotesize
       \setstretch{1}
       {\fontfamily{pcr}\selectfont
\begin{lstlisting}
"""Task prompt (the instruction we want the video to satisfy): "{task_prompt}"

You are a video semantic evaluator. You are given ONE whole generated video.
**Be lenient:** prefer judging the video as consistent with the prompt when the **overall intent**
(what object is manipulated, what action, and the goal) is satisfied. Minor visual differences,
imperfect execution, lighting, or wording-style differences do **not** count as mismatch.

## Step 1 - Does the video fit the prompt? (lenient)
Answer YES if the video **substantially** shows what the prompt asks: same task goal, correct kind of
action, and the right kind of objects/scene - even if details are imperfect.
Answer NO only if there is a **clear** conflict (wrong object, wrong action, wrong goal, or major
contradiction with the prompt). When uncertain, lean toward **YES (fit)**.
For object counts: only treat as mismatch if there is an **obvious** contradiction with the prompt or
with the Prior first-frame scene JSON (not borderline cases).

## Step 2 - Produce modified_prompt
- **If fit (YES):** set modified_prompt to the **exact same string** as the task prompt above
  (verbatim copy; do not rephrase or "improve" wording).
- **If not fit (NO):** make the **smallest possible edit** to the task prompt so the new text matches
  what the video actually shows. Change only the words/phrases that are wrong; keep all other tokens
  identical to the original. Do **not** fully rewrite or paraphrase the whole sentence unless every
  part is wrong. The result should still read like a minimal patch, not a new instruction from scratch.

## Step 3 - Tags
Add tags such as "wrong_object", "wrong_action", "wrong_relation", "wrong_goal", "wrong_count",
"count_mismatch" **only** when you answered NO in Step 1. If you answered YES, use an empty list [].

## Output Format (JSON only, no markdown)
{{
    "reasoning": "Whether the prompt fit the video (lenient); if NO, what was wrong and what minimal edits you applied.",
    "modified_prompt": "...",
    "tags": []
}}
"""
\end{lstlisting}
}
\end{tcolorbox}

\begin{tcolorbox}[notitle, sharp corners, breakable, colframe=teal!40, colback=gray!4, 
       boxrule=3pt, boxsep=0.5pt, enhanced, 
       shadow={3pt}{-3pt}{0pt}{opacity=1,gray!10},
       title={Prompt for Frame-level ($s_F$) Rewarding}]\label{box:prompt}
       \footnotesize
       \setstretch{1}
       {\fontfamily{pcr}\selectfont
\begin{lstlisting}
"""f"Task prompt: {task_prompt}\n\n"
f"You evaluate **frame-level continuity**. Left/first = earlier keyframe, right/second = later keyframe (pair {i}->{i+1}).\n"
"**Scope (only these; ignore deformation/melting/global physics - evaluated elsewhere).**\n"
"**Policy: when uncertain, fail.** `score` must be integer **0 or 1**.\n\n"
f"(1) **Object existence:** For anchors ({anchor_hint}), no **unexplained disappearance** or **replacement/identity swap**.\n"
"(2) **Position jump:** No **unreasonable sudden teleport** of a tracked object between these two frames (large discontinuous jump inconsistent with video framerate/motion).\n"
"(3) **Occlusion / penetration:** No new or worsened **inter-penetration** between objects or environment."
f"{count_block_pair}"
"`tags` may contain **only**: "
f"{tag_list_pair}. Else `[]`.\n\n"
"Output ONLY valid JSON (no markdown): `reasoning`, `score` (0 or 1), `tags` (array)."
"""
\end{lstlisting}
}
\end{tcolorbox}

\begin{tcolorbox}[notitle, sharp corners, breakable, colframe=teal!40, colback=gray!4, 
       boxrule=3pt, boxsep=0.5pt, enhanced, 
       shadow={3pt}{-3pt}{0pt}{opacity=1,gray!10},
       title={Prompt for Segment-level ($s_S$) Rewarding}]\label{box:prompt}
       \footnotesize
       \setstretch{1}
       {\fontfamily{pcr}\selectfont
\begin{lstlisting}
"""f"Task: {task_prompt}\n\n"
f"You are an action-step evaluator. These images are from **segment {i+1}** of a video.\n"
f"**Single question:** Does this segment clearly show sub-action **'{action_str}'** or its **unambiguous result**?\n"
"**Scope:** Only this step. Ignore deformation, global physics, and object counts vs Prior (evaluated in **s_F**).\n"
"**Policy: when uncertain, fail.** `score` integer **0 or 1**.\n\n"
"If missing/failed, use a tag like `missing_step:grasp`, `missing_step:lift`, `missing_step:move`, or `missing_step:place` as appropriate.\n\n"
"Output ONLY valid JSON (no markdown): `reasoning`, `score`, `tags` (array)."
"""
\end{lstlisting}
}
\end{tcolorbox}

\begin{tcolorbox}[notitle, sharp corners, breakable, colframe=teal!40, colback=gray!4, 
       boxrule=3pt, boxsep=0.5pt, enhanced, 
       shadow={3pt}{-3pt}{0pt}{opacity=1,gray!10},
       title={Prompt for Episode-level ($s_E$) Rewarding}]\label{box:prompt}
       \footnotesize
       \setstretch{1}
       {\fontfamily{pcr}\selectfont
\begin{lstlisting}
"""f"Original task prompt (use this verbatim for goal definition): {task_prompt}\n\n"
"You are an **episode-level** robotics evaluator. Input: ONE whole video (time order preserved).\n"
"This is the **only** place to judge **severe / global physics** and **object integrity** (deformation, melting, mesh glitches, etc.). "
"Frame-level (s_F) scoring elsewhere checks keyframe continuity, including instance-count consistency vs the Prior scene JSON when provided.\n\n"
"**Policy: when uncertain on physics, set physics_valid=0.**\n\n"
"(1) **task_complete** (integer 0 or 1): By the end of the video, is the **goal in the original prompt** clearly achieved?\n"
"(2) **physics_valid** (integer 0 or 1): **0** if **at any moment** there are **serious** physical violations, e.g. "
"floating objects, impossible motion, severe inter-penetration, missing parts; **or** gross **deformation / integrity** failure "
"(rubbery warping, liquifying, mesh artifacts, texture smear, object fusion/split, unrealistic scale explosions).\n\n"
"Output ONLY valid JSON (no markdown): `reasoning`, `task_complete`, `physics_valid`, `tags` (use e.g. `deformation`, `shape_distortion`, "
"`mesh_artifact`, etc. when applicable)."
"""
\end{lstlisting}
}
\end{tcolorbox}

\subsection{More Details of Hierarchical Preference Optimization}

As introduced in Section~\S\ref{sec:4.3}, nighttime consolidation of $\mathcal{P}$ leverages a hierarchical preference optimization strategy to extract rich supervisory signals from daytime exploration. To clarify the exact construction of the preference pairs $(\mathbf{c}, \pi^+, \pi^-)$ for the three cognitive dimensions, we detail the data curation process below.

\begin{itemize}[leftmargin=*,itemsep=4pt,parsep=0pt,topsep=4pt]
    \item \textbf{Planning-level ($\mathcal{D}_P$):} This level aims to enhance the planner's logical consistency and reasoning capabilities given a static observation.
    \begin{itemize}[leftmargin=1.2em,itemsep=2pt,parsep=0pt,topsep=2pt]
        \item[\ding{111}] \textit{Context ($\mathbf{c}$):} The initial seed image $I$.
        \item[\ding{111}] \textit{Positive Sample ($\pi^+$):} The consensus plan derived from the majority voting during the daytime GRPO phase. Note that this is represented in natural language rather than discrete atomic action IDs.
        \item[\ding{111}] \textit{Negative Sample ($\pi^-$):} To construct challenging hard negatives, we utilize two distinct sources: (1) the runner-up plan from the majority voting process, which represents a visually plausible but suboptimal alternative; and (2) a corrupted version of the positive plan generated via random clipping, forcing the model to strictly penalize incomplete logical steps.
    \end{itemize}

    \item \textbf{Understanding-level ($\mathcal{D}_U$):} This level improves dynamic visual grounding by teaching the model to correctly describe physical executions.
    \begin{itemize}[leftmargin=1.2em,itemsep=2pt,parsep=0pt,topsep=2pt]
        \item[\ding{111}] \textit{Context ($\mathbf{c}$):} The full video trajectory $V^+$ that successfully passed $\mathcal{S}$ validation and received a high multi-granular reward during daytime.
        \item[\ding{111}] \textit{Positive Sample ($\pi^+$):} The original task prompt that successfully elicited the valid video execution.
        \item[\ding{111}] \textit{Negative Sample ($\pi^-$):} The runner-up plan from the daytime voting. Since the video strictly executes the consensus plan, the runner-up serves as a visually similar but factually incorrect text description, effectively mitigating perceptual hallucinations.
    \end{itemize}

    \item \textbf{Transition-level ($\mathcal{D}_T$):} This level forces the planner to internalize causality and state-transition dynamics without relying on continuous temporal cues.
    \begin{itemize}[leftmargin=1.2em,itemsep=2pt,parsep=0pt,topsep=2pt]
        \item[\ding{111}] \textit{Context ($\mathbf{c}$):} A state pair comprising only the initial and final frames $(f_1, f_T)$ sliced from the highly-rewarded daytime video $V^+$.
        \item[\ding{111}] \textit{Positive Sample ($\pi^+$):} The true executed task prompt that caused this specific state transition.
        \item[\ding{111}] \textit{Negative Sample ($\pi^-$):} The runner-up plan from the majority voting. This acts as a fine-grained hard negative, requiring the model to keenly differentiate between similar underlying intents based solely on the before-and-after physical states.
    \end{itemize}
\end{itemize}

\subsection{More Details of Evaluation Benchmarks}

In this section, we provide detailed configurations and metric definitions for the benchmarks used to evaluate both the Simulator $\mathcal{S}$ and the Planner $\mathcal{P}$.

\vspace{-0.4em}
\paragraph{Simulator Evaluation.}
Since the raw BridgeData V2 \citep{ebert2021bridge} test set primarily comprises single-stage atomic tasks, we systematically construct a multi-level benchmark to rigorously assess compositional generalization. We leverage Gemini-2.5-Pro \citep{comanici2025gemini} to logically compose basic atomic skills into Level-2 (two-stage) and Level-3 (three-stage) instructions. To ensure a balanced evaluation, each difficulty level contains exactly $214$ task prompts. All composed tasks undergo rigorous human verification to guarantee physical feasibility. 
For evaluation metrics, following \citep{zhang2025mind}, we focus on three primary dimensions:
\vspace{-0.4em}
\begin{itemize}[leftmargin=*,itemsep=2pt,parsep=0pt,topsep=2pt]
    \item \textbf{VBench}: We measure general video generation quality across six core dimensions: Aesthetic Quality, Imaging Quality, Temporal Flicker, Motion Smoothness, Subject Consistency, and Background Consistency.
    \item \textbf{Task Success Rate}: We employ Gemini-2.5-Pro as an automated evaluator. Beyond assessing the entire video holistically, we adopt a granular, step-wise evaluation protocol for composite tasks (Level-2 and Level-3). Partial rewards are granted based on the completion of sub-tasks (\textit{e.g.}, executing only one out of two sub-tasks yields a $50\%$ success rate). An example evaluation prompt for Level-2 is shown here.
\vspace{-0.7em}
\begin{tcolorbox}[notitle, sharp corners, breakable, colframe=MidnightBlue!80, colback=gray!4, 
       boxrule=3pt, boxsep=0.5pt, enhanced, 
       shadow={3pt}{-3pt}{0pt}{opacity=1,gray!10},
       title={Level-2 Evaluation Prompt for Gemini}]\label{box:prompt}
       \footnotesize
       \setstretch{1}
       {\fontfamily{pcr}\selectfont
\begin{lstlisting}
"""You are an expert evaluator for embodied AI and robotic manipulation. Your objective is to assess whether the robotic arm in the provided video successfully executes the given text instruction. 

Task Instruction: "{TASK_PROMPT}"

Evaluation Protocol:
1. Task Decomposition: Break down the given instruction into distinct, sequential atomic sub-tasks (e.g., "pick up the red bowl and place it on the stove" -> Sub-task 1: Pick up the red bowl; Sub-task 2: Place it on the stove).
2. Physical Verification: Carefully observe the video to verify if each sub-task is completely and physically achieved. Pay close attention to object interactions, spatial relations, and physical state changes. 
3. Granular Scoring: Evaluate the video strictly. Award partial credit based on the number of successfully completed sub-tasks. If the robot fails a sub-task, all subsequent sub-tasks that depend on it should also be considered failed.

Output your evaluation STRICTLY in the following JSON format:
{
  "sub_tasks_identified": [
    "1. [First atomic action]",
    "2. [Second atomic action]"
  ],
  "reasoning": "Provide a concise, step-by-step analysis of what the robot actually does in the video. Explicitly state which sub-tasks succeeded and where it failed (e.g., missed grasp, dropped object, wrong target).",
  "total_sub_tasks": <int>,
  "completed_sub_tasks": <int>,
  "success_rate": <0.0, 0.5, or 1.0>
}

"""
\end{lstlisting}
}
\end{tcolorbox}
\vspace{-0.6em}
    \item \textbf{User Preference}: To comprehensively evaluate overall semantic and physical alignment, we conduct a blind user study. We randomly sample $50$ generated videos per difficulty level across all eight evaluated methods (Table~\ref{tab:table1}). Ten human evaluators with expertise in generative models are tasked to independently select the single best execution for each task group.
\end{itemize}

\vspace{-0.4em}
\paragraph{Planner Evaluation.}
We benchmark the reasoning and planning capabilities of $\mathcal{P}$ using EB-ALFRED and EB-Habitat from the widely-adopted EmbodiedBench \citep{yang2025embodiedbench}.
\vspace{-0.4em}
\begin{itemize}[leftmargin=*,itemsep=2pt,parsep=0pt,topsep=2pt]
    \item \textbf{EB-ALFRED}: Built upon the AI2-THOR environment, this benchmark focuses on object-centric household tasks involving logical state changes (\textit{e.g.}, washing, heating). It evaluates the mapping of natural language prompts to semantic macro-actions across $300$ total episodes. These episodes are evenly distributed ($50$ each) across six orthogonal capability dimensions: Base, Common Sense, Complex Instruction, Spatial Awareness, Visual Appearance, and Long Horizon.
    \item \textbf{EB-Habitat}: Built upon the Habitat simulator using photorealistic 3D scans (\textit{e.g.}, Matterport3D, Replica), this benchmark focuses on large-scale spatial navigation (ObjectNav) and cross-room object rearrangement. It rigorously tests 3D spatial awareness, long-term memory, and dynamic replanning in visually complex, occluded environments.
    \item \textbf{Metrics}: For both environments, the primary metric is the \textit{Task Success Rate (SR)}. Unlike the simulator evaluation, this is a strict binary metric. An episode is recorded as successful ($1$) if and only if the final physical state perfectly matches the ground truth. No partial rewards are awarded, ensuring a rigorous assessment of end-to-end planning accuracy.
\end{itemize}
\vspace{-0.4em}


\section{More Results \& Sensitivity Analysis}
\label{app_results}

\subsection{RL Training Curve}

\begin{figure*}[!h]
\centering
\vspace{-1.2em}
\includegraphics[width=\linewidth]{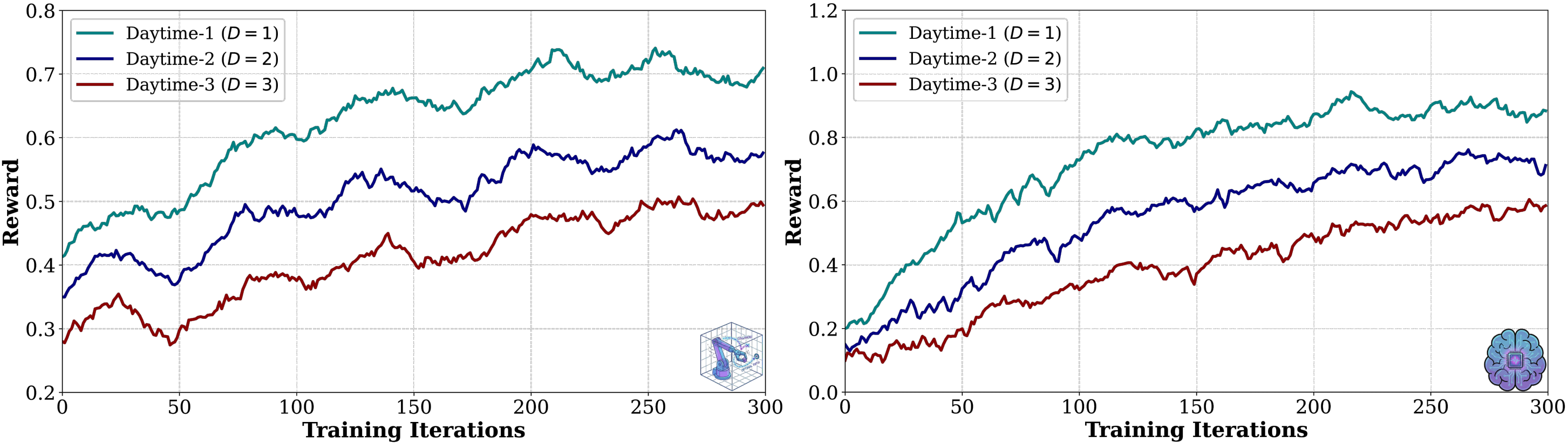}
\vspace{-1.6em}
\caption{Reward curve of (\textbf{\textit{Left}}) Simulator and (\textbf{\textit{Right}}) Planner.}
\label{app_fig:app_Rl}
\vspace{-0.4em}
\end{figure*}

To empirically validate the stability of our online exploration phase, we visualize the Daytime reward curves for both $\mathcal{S}$ and $\mathcal{P}$. Figure~\ref{app_fig:app_Rl} illustrates the training dynamics across three continuous daytime iterations, corresponding to the progressive difficulty levels $D=1, 2$, and $3$. As shown, despite the escalating task complexity and the highly combinatorial nature of complex tasks, both models exhibit steady, monotonic reward convergence over $300$ training iterations. The absence of reward hacking or catastrophic policy collapse effectively demonstrates that our semantic-controlled multi-granular reward provides robust, well-shaped supervisory signals for continuous self-evolution.

\subsection{Analysis of Semantic-Controlled Multi-Granular Reward}

In this section, we investigate the individual contributions of each component within our semantic-controlled multi-granular reward design, as detailed in Table~\ref{app_tab:reward}. Note that to strictly isolate the impact of daytime reward shaping, omitting a specific component in this ablation implies it is excluded from the total daytime reward calculation, though it is still computed in the background to construct "near-miss" preference pairs for nighttime consolidation. 

\begingroup
\setlength{\tabcolsep}{11.4pt}
\begin{table*}[!h]
\vspace{-0.6em}
\renewcommand{\arraystretch}{1.2}
  \centering
  \caption{Ablation study of semantic-controlled multi-granular reward.}
  \centering
  \vspace{-0.7em}
  \footnotesize
   \begin{tabular}{lccccc}
     \hlineB{2.5}
     \rowcolor{CadetBlue!20} 
     \textbf{Rewarding} & \textbf{Level-1$\uparrow$} & \textbf{Level-2$\uparrow$} & \textbf{Level-3$\uparrow$} & \textbf{EB-ALFRED$\uparrow$} & \textbf{EB-Habitat$\uparrow$} \\
     \hlineB{1.5}
     w/o Semantic-alignment $\mathbb{I}_\text{sem}$ & 0.593 & 0.526 & 0.441 & 50.7 & 42.7  \\
     \rowcolor{lightgray!20} 
     w/o Frame-level $s_F$ & 0.631 & 0.544 & 0.452 & 54.3 & 47.7  \\
     w/o Segment-level $s_S$ & 0.626 & 0.549 & 0.458 & 55.3 & 47.3  \\
     \rowcolor{lightgray!20} 
     w/o Episode-level $s_E$ & 0.617 & 0.535 & 0.449 & 53.3 & 46.0  \\
     \textbf{Ours} & \textbf{0.668} & \textbf{0.591} & \textbf{0.505} & \textbf{61.7} & \textbf{54.3}  \\
     \hlineB{2.5}
   \end{tabular}
  \label{app_tab:reward}
  \vspace{-1em}
\end{table*} 
\endgroup

The results in Table~\ref{app_tab:reward} demonstrate that the semantic-alignment indicator ($\mathbb{I}_{sem}$) is arguably the most critical factor. Removing it leads to the most severe performance degradation across all metrics, precipitating a steep drop of $11.0$ and $11.6$ absolute points on the reasoning-heavy EB-ALFRED and EB-Habitat benchmarks. This confirms its indispensable role as a gating mechanism that prevents the system from blindly optimizing physical realism at the expense of semantic task fidelity. Furthermore, omitting any physical granularity ($s_F$, $s_S$, or $s_E$) consistently impairs performance, with the episode-level task success ($s_E$) showing the most pronounced impact among the physical checks. Ultimately, the full variant achieves the highest performance, validating that the tight integration of semantic control and multi-level physical verification is essential for stable co-evolution.

\subsection{Analysis of Selective Simulation Strategy}

In this section, we further evaluate the efficacy of the selective simulation strategy during the daytime exploration of Planner $\mathcal{P}$, with results on the EB-ALFRED benchmark presented in Table~\ref{app_tab:selective}. 

\begingroup
\setlength{\tabcolsep}{8.6pt}
\begin{table*}[!h]
\vspace{-0.6em}
\renewcommand{\arraystretch}{1.2}
  \centering
  \caption{Ablation study of selective simulation strategy.}
  \centering
  \vspace{-0.7em}
  \footnotesize
   \begin{tabular}{lccccccc}
     \hlineB{2.5}
     \rowcolor{CadetBlue!20} 
     \textbf{Simulation Strategy} & \textbf{Base$\uparrow$} & \textbf{Common$\uparrow$} & \textbf{Complex$\uparrow$} & \textbf{Visual$\uparrow$} & \textbf{Spatial$\uparrow$} & \textbf{Long$\uparrow$} & \textbf{Average$\uparrow$} \\
     \hlineB{1.5}
     w/o Selective Simulation & 58 & 48 & 64 & 52 & 48 & 64 & 55.7  \\
     \rowcolor{lightgray!20} 
     \textbf{Ours} & \textbf{60} & \textbf{52} & \textbf{70} & \textbf{62} & \textbf{52} & \textbf{74} & \textbf{61.7}  \\
     \hlineB{2.5}
   \end{tabular}
  \label{app_tab:selective}
  \vspace{-1em}
\end{table*} 
\endgroup

Removing this strategy leads to a distinct performance degradation, dropping the average task success rate by $6.0$ absolute points. Notably, the decline is most pronounced in highly demanding scenarios, such as the \textit{Visual} and \textit{Long} capability dimensions. This degradation suggests that without the self-consistency filtering provided by selective simulation, $\mathcal{P}$ risks over-fitting to or internalizing $\mathcal{S}$'s occasional physical hallucinations during exhaustive rollouts. Furthermore, beyond these empirical performance drops, attempting to physically simulate every generated candidate plan for the Daytime would inevitably introduce prohibitive computational overhead and training time consumption.

\subsection{Analysis of Hierarchical Preference Optimization}

In this section, we further investigate the individual contributions of the three cognitive dimensions within our hierarchical preference optimization strategy during the nighttime consolidation phase. 

\begingroup
\setlength{\tabcolsep}{7.8pt}
\begin{table*}[!h]
\vspace{-0.6em}
\renewcommand{\arraystretch}{1.2}
  \centering
  \caption{Ablation study of hierarchical preference optimization.}
  \centering
  \vspace{-0.7em}
  \footnotesize
   \begin{tabular}{lccccccc}
     \hlineB{2.5}
     \rowcolor{CadetBlue!20} 
     \textbf{Preference Level} & \textbf{Base$\uparrow$} & \textbf{Common$\uparrow$} & \textbf{Complex$\uparrow$} & \textbf{Visual$\uparrow$} & \textbf{Spatial$\uparrow$} & \textbf{Long$\uparrow$} & \textbf{Average$\uparrow$} \\
     \hlineB{1.5}
     w/o Planning-level $\mathcal{D}_P$ & 54 & 44 & 64 & 56 & 46 & 64 & 54.7  \\
     \rowcolor{lightgray!20} 
     w/o Understanding-level $\mathcal{D}_U$ & 56 & 42 & 66 & 52 & 44 & 68 & 54.7  \\
     w/o Transition-level $\mathcal{D}_T$ & 60 & 48 & 66 & 56 & 48 & 68 & 57.7  \\
     \rowcolor{lightgray!20} 
     \textbf{Ours} & \textbf{60} & \textbf{52} & \textbf{70} & \textbf{62} & \textbf{52} & \textbf{74} & \textbf{61.7}  \\
     \hlineB{2.5}
   \end{tabular}
  \label{app_tab:HPO}
  \vspace{-1em}
\end{table*} 
\endgroup

As shown in Table~\ref{app_tab:HPO}, ablating any preference level strictly degrades $\mathcal{P}$'s performance on the EB-ALFRED benchmark, confirming that these hierarchical signals provide orthogonal and synergistic benefits. Notably, both the planning-level ($\mathcal{D}_P$) and understanding-level ($\mathcal{D}_U$) optimizations prove to be particularly foundational; omitting either precipitates a severe $7.0$-point drop in the average success rate. Specifically, without $\mathcal{D}_P$, the model struggles to maintain logical consistency over extended horizons, evidenced by a distinct $10$-point decline in the \textit{Long} capability dimension. Conversely, removing $\mathcal{D}_U$ heavily impairs visual grounding, causing a corresponding $10$-point drop in the \textit{Visual} and \textit{Common} dimensions, indicating a failure to accurately bind textual goals to physical observations. While the transition-level ($\mathcal{D}_T$) ablation exhibits a slightly milder overall degradation, it still noticeably impairs \textit{Complex} and \textit{Visual} reasoning, underscoring its essential role in helping $\mathcal{P}$ internalize state-transition causality.

\subsection{Analysis of Curriculum Hyperparameter}

In this section, we analyze the sensitivity of the curriculum exploration hyperparameter $\lambda$ in Equation~(\ref{eq:curriculum}), which governs the pacing of difficulty progression. 

\begingroup
\setlength{\tabcolsep}{1.6pt}
\begin{table*}[!h]
\vspace{-0.6em}
\renewcommand{\arraystretch}{1.2}
  \centering
  \caption{Ablation study of curriculum hyperparameter.}
  \centering
  \vspace{-0.7em}
  \footnotesize
   \begin{tabular}{lccccccccc}
     \hlineB{2.5}
     \rowcolor{CadetBlue!20} 
     \textbf{$\lambda$} & \textbf{Daytime-1} & \textbf{Nighttime-1} & \textbf{Daytime-2} & \textbf{Nighttime-2} & \textbf{Daytime-3} & \textbf{Nighttime-3} &  \textbf{Level-1$\uparrow$} &  \textbf{Level-2$\uparrow$} &  \textbf{Level-3$\uparrow$}\\
     \hlineB{1.5}
     0.01 & $D=1$ & $D=1$ & $D=1$ & $D=1$ & $D=1$ & $D=1$ & 0.621 & 0.484 & 0.396  \\
     \rowcolor{lightgray!20} 
     0.05 & $D=1$ & $D=1$ & $D=1$ & $D=1$ & $D=2$ & $D=2$ & 0.654 & 0.535 & 0.439 \\
     0.10 \textbf{(Ours)} & $D=1$ & $D=1$ & $D=2$ & $D=2$ & $D=3$ & $D=3$ & \textbf{0.668} & \textbf{0.591} & \textbf{0.505}  \\
     \hlineB{2.5}
   \end{tabular}
  \label{app_tab:curriculum}
  \vspace{-1em}
\end{table*} 
\endgroup

As shown in Table~\ref{app_tab:curriculum}, setting $\lambda$ to a highly conservative value (\textit{e.g.}, $\lambda = 0.01$) effectively paralyzes the curriculum, trapping the system in $D=1$ for all three dual-phase iterations. Crucially, we observe a counter-intuitive phenomenon: this prolonged, exclusive training on simple atomic tasks yields strictly inferior performance even on Level-1 evaluation ($0.621$) compared to our progressive setting ($0.668$). This indicates that endlessly exploiting a saturated difficulty traps the models in local optima and wastes computational budget on marginal, redundant updates. A moderate value ($\lambda = 0.05$) slightly accelerates learning but still delays the transition to complex tasks. In contrast, our optimal setting ($\lambda = 0.10$) organically maps one day-night cycle to one difficulty ascension ($D=1 \rightarrow 2 \rightarrow 3$). This appropriate scaling not only efficiently unlocks multi-stage capabilities but also induces a positive backward transfer, where mastering higher-level compositional tasks inherently reinforces and elevates the robustness of foundational atomic skills.

\subsection{Physical Commonsense Evaluation of Simulator}
\label{app:videophy}

To further assess the semantic and physical properties of the evolved video simulator, we evaluate it on VideoPhy-1~\citep{bansal2025videophy} and VideoPhy-2~\citep{bansal2026videophy}. These benchmarks provide externally specified prompts and assess semantic adherence (SA) and physical commonsense (PC). We use VideoCon-Physics for VideoPhy-1 and VideoPhy2-AutoEval for VideoPhy-2 to obtain automated scores. This experiment evaluates the simulator trained in the VLM-VGM setting; the external evaluators are used for assessment.

\begingroup
\setlength{\tabcolsep}{2.4pt}
\begin{table*}[!h]
\vspace{-0.6em}
\renewcommand{\arraystretch}{1.2}
  \centering
  \caption{Evaluation of \ourmethod's Simulator $\mathcal{S}$ on VideoPhy-1 and VideoPhy-2.}
  \centering
  \vspace{-0.7em}
  \footnotesize
   \begin{tabular}{lcccc}
     \hlineB{2.5}
     \rowcolor{CadetBlue!20} 
     \textbf{Method} & \textbf{VideoPhy-1-SA$\uparrow$} & \textbf{VideoPhy-1-PC$\uparrow$} & \textbf{VideoPhy-2-SA$\uparrow$} & \textbf{VideoPhy-2-PC$\uparrow$} \\
     \hlineB{1.5}
     Cosmos-Diffusion-7B \citep{agarwal2025cosmos} & 57.0 & 18.0 & 26.3 & 54.2   \\
     CogVideoX-5B \citep{yang2024cogvideox} & 63.1 & 31.4 & 28.9 & 68.4   \\
     VideoREPA \citep{zhang2025videorepa} & 51.9 & 22.4 & 21.0 & 72.5   \\
     \hline
     Wan2.2 \citep{wan2025wan} & 41.5 & 25.2 & 24.5 & 69.2  \\
     \rowcolor{lightgray!20} 
     \textbf{+ \ourmethod (Ours)} & \textbf{64.2} & \textbf{32.3} & \textbf{30.2} & \textbf{82.7} \\
     \hlineB{2.5}
   \end{tabular}
  \label{app_tab:videophy}
  \vspace{-1em}
\end{table*} 
\endgroup

Table~\ref{app_tab:videophy} shows consistent improvements over the base Wan2.2 model on both benchmarks. On VideoPhy-1, SA increases from $41.5$ to $64.2$ and PC from $25.2$ to $32.3$. On VideoPhy-2, SA increases from $24.5$ to $30.2$ and PC from $69.2$ to $82.7$. The PC improvements of $7.1$ and $13.5$ score points complement the manipulation-specific evaluation in the main text, showing that the evolved simulator also obtains stronger physical commonsense scores under external evaluation. The simultaneous SA gains indicate improved alignment with the benchmarks' externally specified instructions.

\subsection{Effect of the Base Planner}
\label{app:base_planner}

To examine how the choice of planner affects co-evolution, we instantiate \ourmethod with InternVL3-8B \citep{zhu2025internvl3} and compare it with the default Qwen3-VL-4B \citep{bai2023qwen} planner. InternVL3 exhibits lower initial performance on EB-ALFRED, with an average success rate of $10.7\%$ compared to Qwen3-VL's $25.3\%$. We evaluate both the resulting simulator performance and the planner's improvement to assess whether co-evolution remains effective under this initialization.

\begingroup
\setlength{\tabcolsep}{31.4pt}
\begin{table*}[!h]
\vspace{-0.6em}
\renewcommand{\arraystretch}{1.2}
  \centering
  \caption{Simulator performance with different base planners.}
  \centering
  \vspace{-0.7em}
  \footnotesize
   \begin{tabular}{lccc}
     \hlineB{2.5}
     \rowcolor{CadetBlue!20} 
     \textbf{Method} & \textbf{Level-1$\uparrow$} & \textbf{Level-2$\uparrow$} & \textbf{Level-3$\uparrow$}\\
     \hlineB{1.5}
     Wan2.2 \citep{wan2025wan} & 0.477 & 0.395 & 0.324  \\
     \rowcolor{lightgray!20} 
     + SFT (Cold Start) & 0.491 & 0.409 & 0.341  \\
     \textbf{+ \ourmethod (InternVL3)} & 0.645 & 0.565 & 0.489  \\
     \rowcolor{lightgray!20} 
     \textbf{+ \ourmethod (Qwen3-VL)} & 0.668 & 0.591 & 0.505  \\
     \hlineB{2.5}
   \end{tabular}
  \label{app_tab:planner_simulator}
\end{table*} 
\endgroup

\begingroup
\setlength{\tabcolsep}{7.8pt}
\begin{table*}[!h]
\renewcommand{\arraystretch}{1.2}
  \centering
  \caption{Planner task success rates (\%) on EB-ALFRED with different base models.}
  \centering
  \vspace{-0.7em}
  \footnotesize
   \begin{tabular}{lccccccc}
     \hlineB{2.5}
     \rowcolor{CadetBlue!20} 
     \textbf{Method} & \textbf{Base$\uparrow$} & \textbf{Common$\uparrow$} & \textbf{Complex$\uparrow$} & \textbf{Visual$\uparrow$} & \textbf{Spatial$\uparrow$} & \textbf{Long$\uparrow$} & \textbf{Average$\uparrow$} \\
     \hlineB{1.5}
     InternVL3 \citep{zhu2025internvl3} & 20 & 12 & 20 & 8 & 2 & 2 & 10.7 \\
     \rowcolor{lightgray!20} 
     \textbf{+ \ourmethod} & \textbf{56} & \textbf{44} & \textbf{58} & \textbf{50} & \textbf{42} & \textbf{56} & \textbf{51.0}  \\
     \hline
     Qwen3-VL \citep{bai2023qwen} & 28 & 20 & 26 & 32 & 20 & 26 & 25.3 \\
     \rowcolor{lightgray!20} 
     \textbf{+ \ourmethod} & \textbf{60} & \textbf{52} & \textbf{70} & \textbf{62} & \textbf{52} & \textbf{74} & \textbf{61.7}  \\
     \hlineB{2.5}
   \end{tabular}
  \label{app_tab:planner_backbone}
  \vspace{-1em}
\end{table*} 
\endgroup

As shown in Table~\ref{app_tab:planner_simulator}, using InternVL3 still improves the simulator across all three difficulty levels, including an increase from $0.341$ to $0.489$ on Level-3 over the cold-started baseline. Meanwhile, Table~\ref{app_tab:planner_backbone} shows that InternVL3's average planning success rises from $10.7\%$ to $51.0\%$, a gain of $40.3$ absolute points, with improvements across all six capability dimensions. Although Qwen3-VL achieves higher final performance, both initializations yield substantial gains in the planner and simulator. These results support the effectiveness of \ourmethod across planner backbones with different initial planning capabilities.

\section{Exhibition Board}

\begin{figure*}[!t]
\centering
\includegraphics[width=\linewidth]{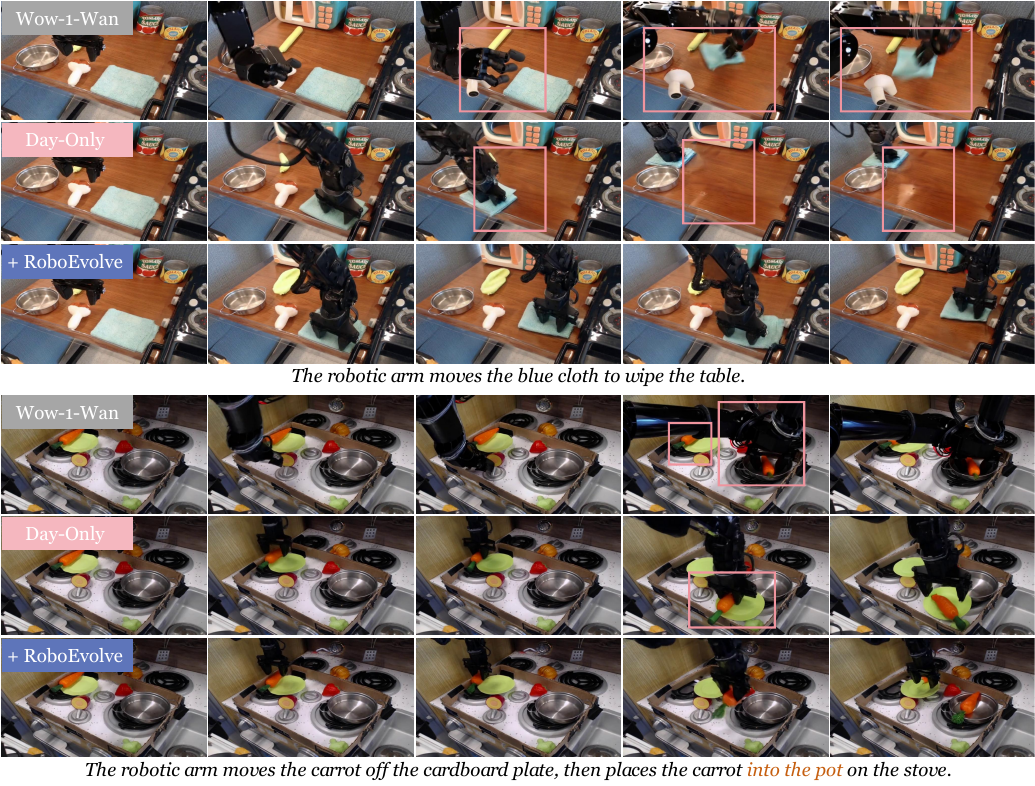}
\vspace{-1.6em}
\caption{More comparison demonstrations of \ourmethod.}
\label{app_fig:app_quali1}
\vspace{-1em}
\end{figure*}

\begin{figure*}[!t]
\centering
\vspace{-1em}
\includegraphics[width=0.9\linewidth]{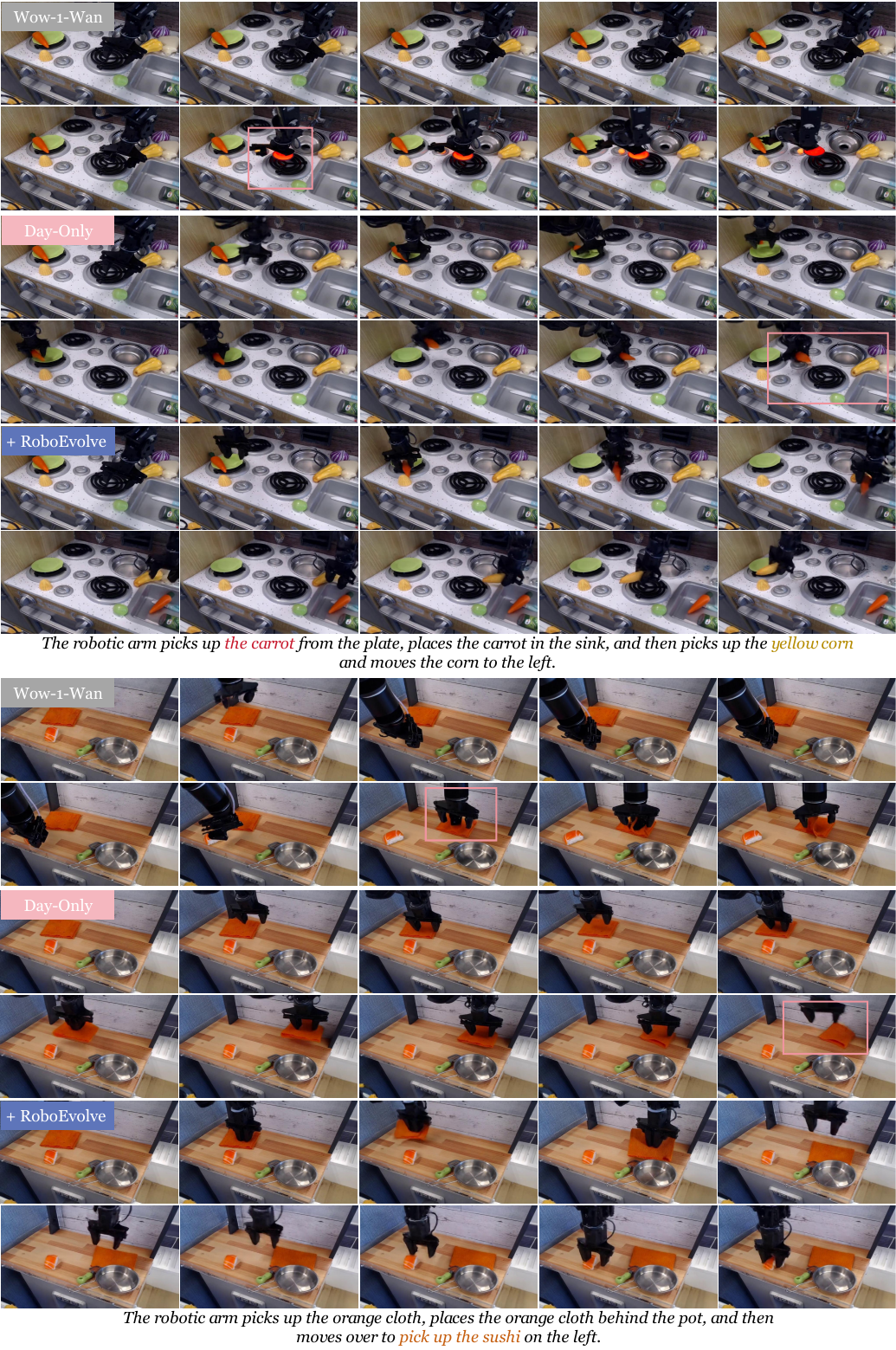}
\vspace{-0.6em}
\caption{More comparison demonstrations of \ourmethod.}
\label{app_fig:app_quali2}
\vspace{-1.3em}
\end{figure*}

\begin{figure*}[!t]
\centering
\vspace{-1em}
\includegraphics[width=0.9\linewidth]{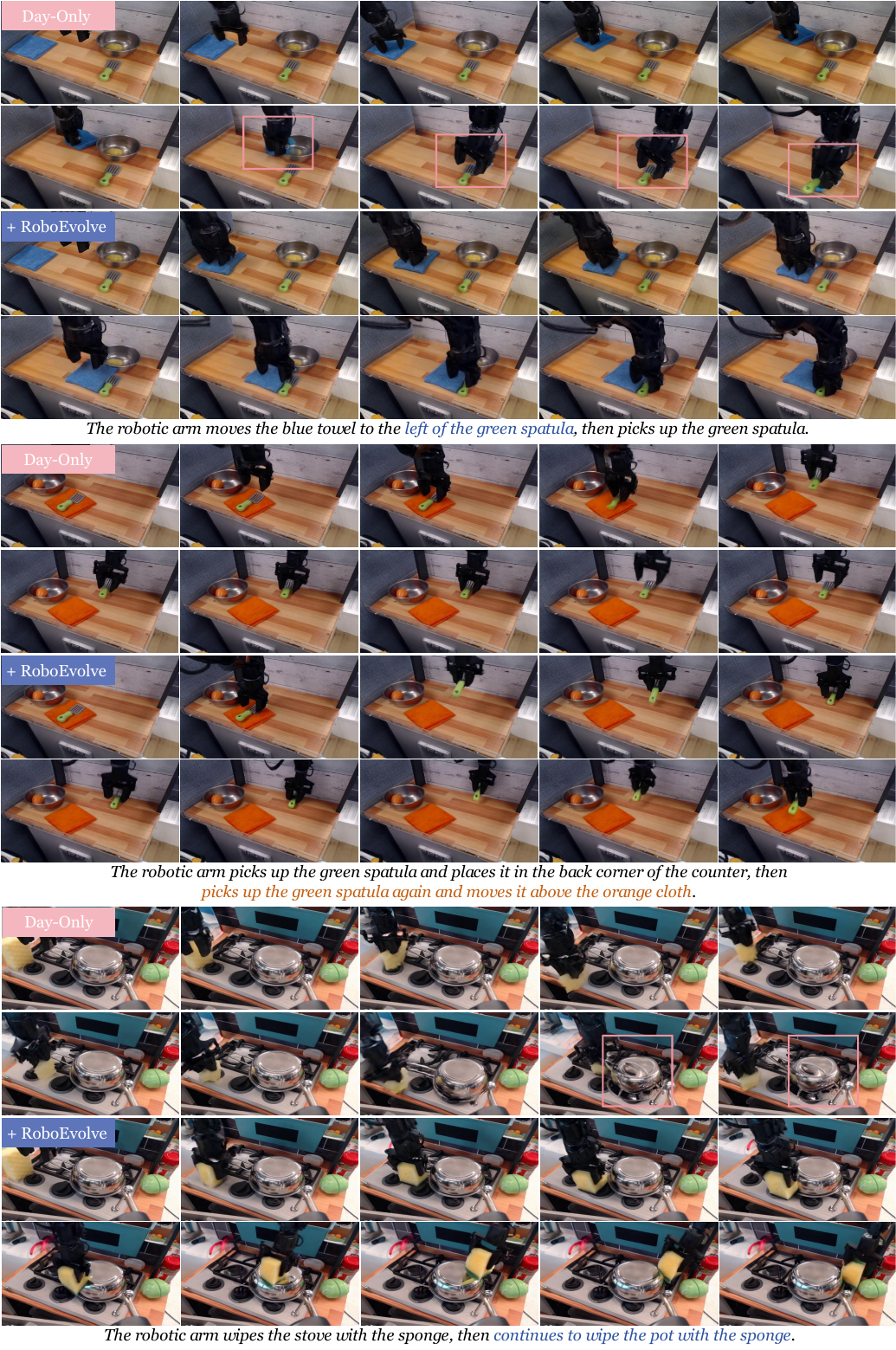}
\vspace{-0.6em}
\caption{More comparison demonstrations of \ourmethod.}
\label{app_fig:app_quali3}
\vspace{-1.3em}
\end{figure*}

In this section, we provide extended qualitative comparisons to further illustrate the differences between \ourmethod and the \textit{Daytime-Only} baseline (Figures~\ref{app_fig:app_quali1}, \ref{app_fig:app_quali2}, and \ref{app_fig:app_quali3}). 

As highlighted in our main analysis, while the Daytime-Only variant is capable of initiating complex, multi-step instructions, it is severely prone to temporal and physical hallucinations. A recurrent failure mode observed in these comparisons is the sudden disappearance of manipulated objects or tools during the transition between sub-tasks (indicated by the pink bounding boxes). This visually demonstrates that without the critical nighttime consolidation phase to learn from "near-miss" negative samples, the model fails to internalize fundamental physical laws such as object permanence. 



\section{Limitations and Future Work}
\label{app_sec:limitation}

While \ourmethod establishes a highly autonomous, self-improving paradigm for robotic manipulation, it possesses certain limitations that present exciting avenues for future research.

First, although our LingBot-VA instantiation extends \ourmethod to low-level action execution in LIBERO, we have not yet deployed the evolved models on physical robotic hardware. This instantiation uses feedback from the simulation environment, and transferring the learned capabilities to real robots remains an important next step. Future work will investigate robustness to differences in sensing, contact dynamics, and control between simulation and physical deployment. Second, in the VLM-VGM instantiation, \ourmethod relies on the zero-shot evaluation capabilities of the planner to assess multi-granular physics and semantics. The VideoPhy results provide additional evidence of improved generated-video quality, while the reliability of these internal judgments remains dependent on the base VLM.



\section{Ethical Implications}
\label{app_sec:ethical}

While \ourmethod significantly advances the data efficiency and scalability of embodied AI without relying on extensive human annotations, its autonomous, self-improving nature introduces relevant ethical considerations. Specifically, because the co-evolutionary loop relies heavily on the zero-shot priors of base VLMs, any latent biases inherent in these foundation models, such as cultural or demographic skewness in object affordances and household environments, could be inadvertently amplified during the synthetic data generation and preference optimization phases. Furthermore, as this paradigm paves the way for highly autonomous physical agents, ensuring rigorous operational safety, value alignment, and human oversight becomes paramount. We encourage the community to proactively integrate safety guardrails and comprehensive fairness evaluations as these closed-loop, self-evolving systems transition from generative simulations to real-world deployments.

\end{document}